\def\year{2019}\relax
\documentclass[letterpaper]{article} %DO NOT CHANGE THIS
\usepackage{aaai20}  %Required
\usepackage{times}  %Required
\usepackage{helvet}  %Required
\usepackage{courier}  %Required
\usepackage{url}  %Required
\usepackage{graphicx}  %Required

\usepackage{hyperref}       % hyperlinks
\usepackage{url}            % simple URL typesetting
\usepackage{booktabs}       % professional-quality tables
\usepackage{amsfonts}       % blackboard math symbols
\usepackage{nicefrac}       % compact symbols for 1/2, etc.
\usepackage{microtype}      % microtypography

\usepackage{amsmath}
\usepackage{subfigure}
\graphicspath{ {figure/} }
\usepackage{xcolor}
\usepackage{tabularx}

\usepackage{wrapfig,lipsum}
\usepackage{caption}
\usepackage{float}

\begin{document}

\title{``The Squawk Bot''\thanks{The name of this work is inspired by the popular financial commentary show called ``Squawk Box'' on CNBC  (CNBC. Squawk Box: Business, Politics, Investors and Traders, https://www.cnbc.com/squawk-box-us/ , accessed May 23rd, 2019)}:~Joint Learning of Time Series and Text Data Modalities for Automated Financial Information Filtering}
\author{Xuan-Hong Dang, Syed Yousaf Shah, Petros Zerfos\\
IBM Research\\
Yorktown Heights, New York 10598\\
xuan-hong.dang@ibm.com, \{syshah,pzerfos\}@us.ibm.com
}
%\vspace*{-0.5cm}
\maketitle
\vspace*{-1.2cm}

\begin{abstract}

Multimodal analysis that uses numerical time series and textual corpora as input data sources is becoming a promising approach, especially in the financial industry. However, the main focus of such analysis has been on achieving high prediction accuracy while little effort has been spent on the important task of understanding the association between the two data modalities. Performance on the time series hence receives little explanation though human-understandable textual information is available. In this work, we address the problem of given a numerical time series, and a general corpus of textual stories collected in the same period of the time series, the task is to timely discover a succinct set of textual stories associated with that time series. Towards this goal, we propose a novel multi-modal neural model called \verb"MSIN" that jointly learns both numerical time series and categorical text articles in order to unearth the association between them. Through \text{multiple} steps of data interrelation between the two data modalities, \verb"MSIN" learns to focus on a small subset of text articles that best align with the performance in the time series. This succinct set is timely discovered and presented as recommended documents, acting as automated information filtering, for the given time series. We empirically evaluate the performance of our model on discovering relevant news articles for two stock time series from Apple and Google companies, along with the daily news articles collected from the Thomson Reuters over a period of seven consecutive years. The experimental results demonstrate that \verb"MSIN" achieves up to 84.9\% and 87.2\% in recalling the ground truth articles respectively to the two examined time series, far more superior to state-of-the-art algorithms that rely on conventional attention mechanism in deep learning.
\end{abstract}

% how text information can be used to provide human-friendly explanation of the performance of the time series being modeled. 
%  accuracy~\cite{schumaker12yahoonews,akita2016deep,weng2017stock}

% We address in this study a challenging yet important problem of jointly learn both time series and textual data in order to timely discover a small succinct set of relevant textual clues aligned with the current state (behavior/performance) of a given time series. Solving this problem is desirable to a number of emerging applications. 

\vspace*{-0.2cm}
%=========================================================================%
\section{Introduction}                                   \label{Introduction}
%=========================================================================%

\begin{figure}
\centering
\includegraphics[width=8cm]{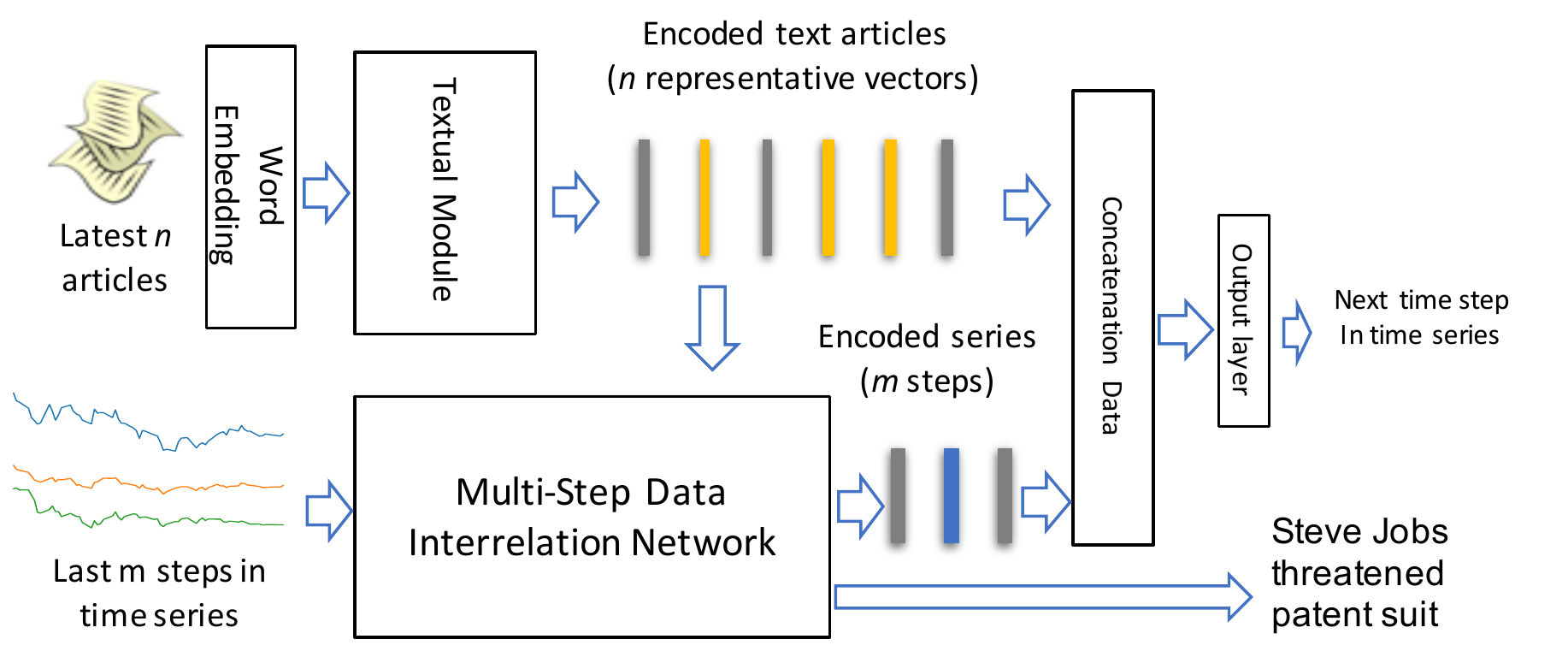}
%\hspace*{-2.5cm}
\caption{Illustration of our model that learns to daily discover top relevant text articles timely associated with the current state (characterized by the last $m$ time steps) in a given time series.}
\vspace*{-0.5cm}
\label{OverallModel}
\end{figure}

Current multimodal analysis that combines time series with text data
%(such as stock prices with financial news and social media trends~\cite{schumaker12yahoonews,akita2016deep,weng2017stock}) 
often focuses on extracting features from text corpus and incorporates them into a forecasting model for enhancing prediction. Little attention is paid to the aspect of using text as a means of explaination for the patterns observed in the time series~\cite{akita2016deep,weng2017stock}. In many emerging applications, given a time series, one can ask for finding a small set of documents that can reflect or influence the time series. Taking ``quantamental investing''~\cite{quantamentalFT} as an example. When trading a stock, investors do not solely base their decisions on its historical prices. Rather, the decisions are made with a careful consideration of the news and events collected from the markets. With the dramatically increasing amount of available news nowadays~\cite{schumaker12yahoonews}, a natural question hence to ask is what would be the most relevant news associated with a particular stock series. As illustrated in Figure~\ref{OverallModel}, given recent historical values of Apple stock time series as one input, and today's textual news collected from a public media as another input, ``Steve Jobs threatened patent suit" is outputted as relevant news. 
Certainly, for different stock time series, the set of relevant associated news articles would be different. Likewise with the cloud business, accurately associating a cloud monitoring metric (time series) with textual complaints from clients can help technicians focus on the right issues, tremendously reducing their efforts and expertise in resolving complaints. Though filtering relevant text stories can rely on keywords, we argue that accurately identifying such keywords associated with a specific time series is not an easy task as requiring domain knowledge. Moreover, in applications like cloud systems, it is unclear on what would be keywords associated with each of thousands of monitoring time series. Furthermore, filtering information based on \textit{fixed} keywords would potentially lose essential information as time series keeps changing and so do the textual contents. Hence an automated system is highly desirable and becomes more relevant.

%\vspace*{-0.3cm}

%  The strict alignment of time steps or sampling frequencies between the two data modalities is not required, as long as we believe that the collected text articles may contain information that reflects the current state of time series or influences its performance in the next time step $t+1$. For example $n$ news articles can be collected from the same period from $t-m$ to $t$ of the time series sequence, or just from the latest day $t$.
 
In this paper, we address the above mentioned important, yet challenging problem through developing a novel, multi-modal, neural network that jointly learns numerical time series and textual documents in order to discover the relation between them. The discovered text articles are returned as a means of recommended documents for the current state of the time series. As shown in Fig.\ref{OverallModel}, our model consists of (1) a textual module that learns representative vectors for input text documents, (2) the \verb"MSIN" (Multi-Step Interrelation Network), which takes as input both the time series and the sequence of textual representative vectors to learn the association between them and hence discovering relevant text documents, (3) a concatenation and dense layer that aggregates information from the two data modalities to output the next value in the time series.

The novelty from our proposed model stems from the introduction of the Multi-Step Interrelation Network, which allows the incorporation of semantic information learnt from the textual modality to every time step modeling of the time series. We argue that \textit{multiple data interrelations} are important and necessary in order to discover the association between the \textit{categorical} text and \textit{numerical} time series. This is because not only they are of different data types, but also no prior knowledge is given on guiding the machine learning model to look at what text in relation to a given time series. Hence, compared to existing multimodal approaches either learning two data modalities sequentially or in parallel, our proposed \verb"MSIN" network effectively leverages the mutual information impact between the two data modalities, textual representative vectors (learnt by the Text Module in Fig.\ref{OverallModel}) and time series' values, through multiple steps of data convolution. During such process, it gradually assigns large attention weights to ``important" text vectors, while rules out less relevant ones, and finally converges to an attention distribution over the text articles that best aligns with the current state of the time series. \verb"MSIN" also technically relaxes the strict alignment between the two data modalities (e.g. at the time stamp level), allowing it to concurrently deal with two (or multiple) input data sequences of different lengths, which currently is not supported by most conventional recurrent neural networks ~\cite{hochreiter1997long,cho2014learning,chung2014empirical}. 
We perform detailed empirical analysis over large-scale financial news and stock prices datasets spanning over seven consecutive years. The model was trained on data of the first six years and evaluated on the last year. We show that \verb"MSIN" achieves strong performance of 84.9\% \& 87.2\% in recalling ground-truth relevant news w.r.t. two specific stock time series, the Apple (AAPL) and Google (GOOG), which are prominent companies during the examined period.

\section{Learning text articles representation}\label{text_module}
%We address the above problem through a novel neural network architecture shown in Fig.~\ref{OverallModel}. 

The first network component in our model is the \textit{textual module} that learns to represent text articles as numerical vectors so that they are comparable to the numerical time series. The order among words within each text article is important in learning their semantic dependencies. Hence, our network exploits the long-short term memory (LSTM)~\cite{hochreiter1997long} to learn such dependencies and aggregates them into an article's representative vector.

At each time stamp $t$, an input data sample to the textual module is a sequence of $n$ text documents $\{doc_1,doc_2,\ldots, doc_n\}$ (e.g., $n$ news articles released at day $t$ by Thomson Reuters). And its output sample is a sequence $n$ representative vectors denoted $\{\mathbf{s}_1,\mathbf{s}_2,\ldots,\mathbf{s}_n\}$ (we omit notation $t$ in these $\mathbf{s}_j$'s and $doc_j$'s to minimize clutter). Each text document $doc_j$ (with $1\leq j\leq n$) in turn is a sequence 
of maximum $K$ words denoted by $doc_j = \{\mathbf{x}^{txt}_1, \mathbf{x}^{txt}_2,\ldots,\mathbf{x}^{txt}_K\}$ (the superscript $txt$ denotes for text modality). Each $\mathbf{x}^{txt}_{\ell} \in \mathbb{R}^V$ is the one-hot-vector representation of the $\ell$-th word in document $doc_j$, with $V$ as the vocabulary size. We use an embedding layer to transform each $\mathbf{x}^{txt}_{\ell}$ into a low dimensional dense vector  $\mathbf{e}_{\ell}\in \mathbb{R}^{d_{w}}$ via a linear transformation: 

\vspace*{-0.2cm}
\begin{align}
\mathbf{e}_{\ell}= \mathbf{W}_{e}*\mathbf{x}^{txt}_{\ell},  \quad \text{in which} \quad   \mathbf{W}_{e} \in \mathbb{R}^{d_{w}\times V}
\end{align}
\vspace*{-0.2cm}

%$\mathbf{e}_{\ell}= \mathbf{W}_{emb}\mathbf{x}^{txt}_{\ell} \text{, s.t. } \mathbf{W}_{emb} \in \mathbb{R}^{d\times V}$. 

Often, $d_{w}$ is much smaller than $V$, and $\mathbf{W}_{e}$ can be trained from scratch; however, using or initializing it with pre-trained vectors from GloVe~\cite{GLOVE} can produce more stable results. In our examined datasets (see Section \ref{Experiment}), we found that setting $d_{w}=50$ is sufficient given the vocabulary size $V=5000$. 

%
% %\vspace*{-0.3cm}
%\begin{figure}[!h]
% \centering
%\includegraphics[width=9cm]{txtLSTM.pdf}
%\hspace*{-1.5cm}
%\caption{\todo{Fig to be updated }Financial news sequence encoding LSTM.}
%%\vspace*{-0.8cm}
%\label{txtLSTM}
%\end{figure}

The sequence of embedded words $\{\mathbf{e}_1, \mathbf{e}_2,\ldots,\mathbf{e}_K\}$ for a document article is then fed into an LSTM that learns to produce their corresponding encoded contextual vector. The key components of an LSTM unit are the memory cell which preserves essential information of the input sequence through time, and the non-linear gating units that regulate the information flow in and out of the cell. At each step $\ell$ (corresponding to $\ell$-th word) in the input sequence, LSTM takes in the embedding vector $\mathbf{e}_{\ell}$, its previous cell state $\mathbf{c}^{txt}_{\ell-1}$, and the previous output vector $\mathbf{h}^{txt}_{\ell-1}$,  to update the memory cell $\mathbf{c}^{txt}_{\ell}$, and subsequently outputs the hidden representation $\mathbf{h}^{txt}_{\ell}$ for $\mathbf{e}_{\ell}$. From this view, we can briefly represent LSTM as a \textit{recurrent} function $\mathbf{f}$ as follows:

\vspace*{-0.2cm}
\begin{align}
\mathbf{h}^{txt}_{\ell} &= \mathbf{f}(\mathbf{e}_{\ell},\mathbf{c}^{txt}_{\ell-1} \mathbf{h}^{txt}_{\ell-1}) ,  \quad \text{for} \quad \ell = 1,\ldots,K \label{txt_hidd}
\end{align} 
\vspace*{-0.2cm}

\noindent in which the memory cell $\mathbf{c}^{txt}_{\ell}$ is updated internally. Both $\mathbf{c}^{txt}_{\ell}, \mathbf{h}^{txt}_{\ell} \in \mathbb{R}^{d_h}$ in which $d_h$ is the number of hidden neurons. Our implementation of LSTM closely follows the one presented in~\cite{zaremba2014recurrent} with two extensions. First, in order to better exploit the semantic dependencies of an $\ell$-th word with both its preceding and following contexts, we build two LSTMs respectively taking the sequence in the forward and backward directions (denoted by the head arrows in Eq.\eqref{hidd_word}). This results in a bidirectional LSTM (BiLSTM): 

\vspace*{-0.2cm}
\begin{align}
\overrightarrow{\mathbf{h}}^{txt}_{\ell} &= \overrightarrow{\mathbf{f}}(\mathbf{e}_{\ell},\mathbf{c}^{txt}_{\ell-1}, \overrightarrow{\mathbf{h}}^{txt}_{\ell-1}), \overleftarrow{\mathbf{h}}^{txt}_{\ell} = \overleftarrow{\mathbf{f}}(\mathbf{e}_{\ell},\mathbf{c}^{txt}_{\ell-1}, \overrightarrow{\mathbf{h}}^{txt}_{\ell-1}) \nonumber\\
\mathbf{h}^{txt}_{\ell} &= [\overrightarrow{\boldsymbol{h}}^{txt}_{\ell}, \overleftarrow{\mathbf{h}}^{txt}_{\ell}]  \quad \text{for} \quad \ell=1,\dots,K \label{hidd_word}
\end{align}

The concatenated vector $\mathbf{h}^{txt}_{\ell}$ leverages the context surrounding the $\ell$-th word and hence better characterizes its semantic as compared to the embedding vector $\mathbf{e}_{\ell}$ which ignores the local context in the input sequence. Second, we extend the model by exploiting the weighted mean pooling from all vectors $\{\mathbf{h}^{txt}_{1},\mathbf{h}^{txt}_{2},\ldots,\mathbf{h}^{txt}_{K}\}$ to form the overall representation $\mathbf{s}_j$ of the entire $j$-th text document:

\vspace*{-0.2cm}
\begin{align} 
\mathbf{s}_j &= \frac{1}{K} \sum_{\ell} \beta_\ell * \mathbf{h}^{txt}_{\ell} \quad \text{for} \quad \ell = 1,\ldots,K \nonumber \\ 
\text{where}  \quad \beta_\ell &= \frac{\exp(\mathbf{u}^\top \tanh(\mathbf{W}_\beta*\mathbf{h}^{txt}_{\ell} + \mathbf{b}_{\ell}))}{\sum_\ell \exp(\mathbf{u}^\top \tanh(\mathbf{W}_\beta* \mathbf{h}^{txt}_{\ell} + \mathbf{b}_{\ell}))} \label{txt_context_vec}
\end{align} 
\vspace*{-0.3cm}

%
%\vspace*{-0.3cm}
%\begin{align}
%\beta_\ell &= \frac{\exp(\mathbf{u}^\top \tanh(\mathbf{W}_\beta*\mathbf{h}^{txt}_{\ell} + \mathbf{b}_{\ell}))}{\sum_\ell \exp(\mathbf{u}^\top \tanh(\mathbf{W}_\beta* \mathbf{h}^{txt}_{\ell} + \mathbf{b}_{\ell}))}\\
%\mathbf{s}_j &= \frac{1}{K} \sum_{\ell} \beta_\ell * \mathbf{h}^{txt}_{\ell}
%\end{align}
%\vspace*{-0.3cm}

\noindent in which $\mathbf{u} \in \mathbb{R}^{2d_h}, \mathbf{W}_\beta \in \mathbb{R}^{2d_h\times2d_h}$ are respectively referred to as the parameterized context vector and matrix whose values are jointly learnt with the BiLSTM. This pooling technique resembles the successful one presented in~\cite{conneau2017supervised} that learns multiple views over each input sequence. Our implementation simplifies them by adopting only a single view ($\mathbf{u}$ vector) with the assumption that each document (e.g., a news story or article) contains only one topic (relevant to the time series). Note that, similar to convolutional neural networks~\cite{kim2014convolutional}, a max pooling can also be used in replacement for the mean pooling in defining $\mathbf{s}_j$. We, however, attempt to keep the model simple since using max function adds a non-smooth function and thus generally requires more training data to learn a proper transformation. We apply our text module BiLSTM to every text article collected at time period $t$ and its output is a sequence of representative vectors 
$\{\mathbf{s}_1,\mathbf{s}_2,\ldots,\mathbf{s}_n\}$, each corresponds to one text document at input.

\vspace*{-0.1cm}
\section{Multi-step interrelation of data modalities}
\vspace*{0.1cm}

%\todo{attention, probability mass, is done completely in the TS domain, not text domain}
%\todo{text to TS not the way around with meaning. The matrix structure together with the penalization term gives our model a greater capacity to disentangle the latent information from the input sentence. }

%While the sequential steps in our textual modality are arbitrary numbers of news stories released at a time period $t$, those in a stock time series are generally a fixed step number $m$ looking back in its history.  

%Moreover, while complemental information from the text can benefit the learning process over the time series, they are often hidden by much of other irrelevant information. 

%  Given the information learnt in the textual component, we propose a novel network named \verb"MSIN" to model the time series while ex

% with complemental information learnt from the textual domain, 

%However, time series patterns like trends, variation, temporal lagged or inter-series dependencies~\cite{} are often complex, non-linear, and spreading out entire length of the time series sequence\footnote{We omit interpreting patterns/behaviors learnt from the time series in this work but refer readers to other related ones~\cite{} which recently demonstrate the effectiveness of using neural networks in modeling them}, 

\begin{figure}
\vspace*{-0.2cm}
\centering
%\hspace*{-1.0cm}
\includegraphics[width=9cm]{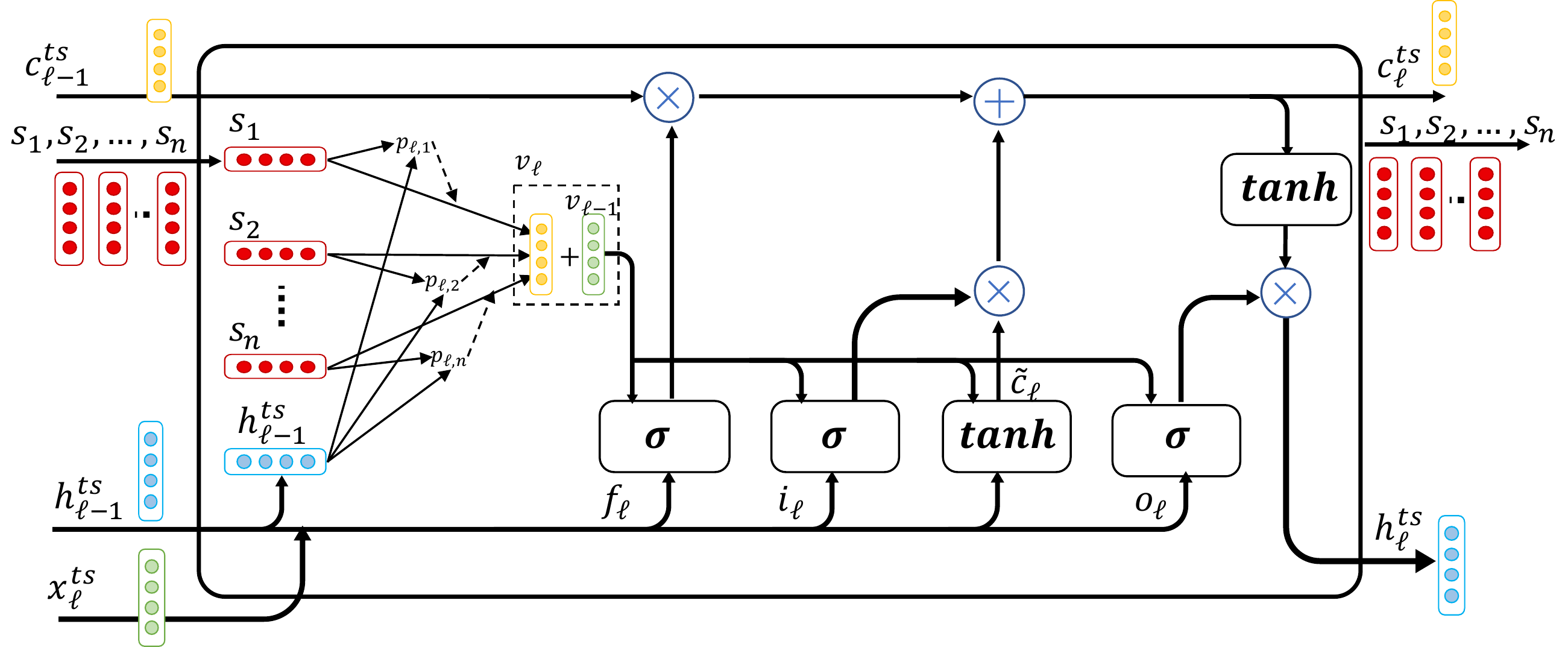}
\vspace*{-0.2cm}
\caption{Memory cell design of the multi-step interrelation network (MSIN).}
\vspace*{-0.4cm}
\label{MSIN}
\end{figure}

%A straightforward approach is to model the time series behaviors independently from the textual news and later align them to find common correlated patterns. However, as time series patterns are often complex, non-linear and especially spread out through the series sequence length, an approach with a single alignment between the final states of two data modalities is often limited in discovering the relevant news in association with the time series. Even more challenging, recall that the input news articles are highly general, noisy and not necessarily reported directly to the input time series as they are collected from the public media. Likewise, there is also no obvious way to synchronize or stepwise-merge the two data modalities since they not only are different in the sequential step lengths but also in the rate of data sampling.

Our next task is to model the time series, taking into account the information learnt from the textual data modality, in order to discover a small set of relevant text articles that well align with the current state of the time series. An approach of using a \textit{single} alignment between the two data modalities (empirically evaluated in Section~\ref{Experiment}) generally does not work effectively since the text articles are highly general, contain noise, and especially we do not have step-wise synchronization between the text and time series. The number of look-back time points $m$ in the time series can be much different from the number of text documents $n$ collected at the time moment, the value that can vary from time to time. To tackle these challenges, we propose the novel \verb"MSIN" network that broadens the scope of LSTM so that it can handle concurrently two input sequences of different lengths. More profoundly, we develop in \verb"MSIN" a novel neural mechanism that integrates information captured in the representative textual vectors learnt in the previous textual module to every step reasoning in the time series sequence. Doing so allows \verb"MSIN" leverage the mutual information between the two data modalities through multi-steps of data interrelation. Consequently, it gradually filters out irrelevant text articles while focuses only on those that correlate with the current patterns learnt from the time series, as it advances in the time series sequence. The chosen text articles are hence captured in the form of a probability mass attended on the input sequence of textual representative vectors. 

In specific, inputs to \verb"MSIN" at each time stamp $t$ are two sequences: (1) values of last $m$ steps in the time series modality $ \{\mathbf{x}^{ts}_{t-m}, \mathbf{x}^{ts}_{t-m-1}, \ldots,\mathbf{x}^{ts}_{t-1}\}$ (superscript $ts$ denotes the time series modality); (2) a sequence of $n$  text representative vectors learnt by the above textual module $\{\mathbf{s}_1,\mathbf{s}_2,\ldots,\mathbf{s}_n\}$. Its outputs are the set of hidden state vectors (described below) and the probability mass vector $p_{m}$ outputted at the last state of the series sequence. The number of entries in $p_{m}$ equal to the number of text vectors at input. Its values are non-negative which encode text documents relevant to the time series sequence. A larger value at $j$-th entry of $p_{m}$ reveals a more relevant $doc_j$ to the time series. 

A memory cell of our \verb"MSIN" is illustrated in Fig.\ref{MSIN}. As observed, the number of gates in \verb"MSIN" remains fixed yet we \textit{augment} the information flow within the cell by the information learnt in the text modality. To be concrete, \verb"MSIN" starts with the initialization of the initial cell state $\mathbf{c}^{ts}_0$ and hidden state $\mathbf{h}^{ts}_0$ by using two separate single-layer neural networks applied on the average state of the text sequence:

%%\vspace*{-0.4cm}
%\begin{align}
%s^{ts}_0 &= \tanh (U_{s0} \sum_i h^{txt}_{i} + b_{s0})\\
%c^{ts}_0 &=\tanh (U_{c0} \sum_i h^{txt}_{i} + b_{c0})
%\end{align}
%%\vspace*{-0.6cm}

%\vspace*{-0.6cm}
%\begin{align}
%\mathbf{c}^{ts}_0 &=\tanh (\mathbf{U}_{c_0} * \overline{\mathbf{s}} + \mathbf{b}_{c_0})\\
%\mathbf{h}^{ts}_0 &= \tanh (\mathbf{U}_{h_0} *  \overline{\mathbf{s}}+ \mathbf{b}_{h_0})
%\end{align}
%\vspace*{-0.4cm}

\vspace*{-0.2cm}
  \begin{align} 
   \label{eq:cts}
 \mathbf{c}^{ts}_0 &=\tanh (\mathbf{U}_{c_0} * \overline{\mathbf{s}} + \mathbf{b}_{c_0})\\
\label{eq:hts} 
\mathbf{h}^{ts}_0 &= \tanh (\mathbf{U}_{h_0} *  \overline{\mathbf{s}}+ \mathbf{b}_{h_0})
\end{align}
\vspace*{-0.2cm}

%
%\vspace*{-0.2cm}
%\noindent\begin{tabularx}{\textwidth}{@{}XX@{}}
%  \begin{align} 
%   \label{eq:cts}
% \mathbf{c}^{ts}_0 =\tanh (\mathbf{U}_{c_0} * \overline{\mathbf{s}} + \mathbf{b}_{c_0})
%\end{align} &
%\begin{align}
%\label{eq:hts} 
%\mathbf{h}^{ts}_0 = \tanh (\mathbf{U}_{h_0} *  \overline{\mathbf{s}}+ \mathbf{b}_{h_0})
%\end{align}
%\end{tabularx}
%\vspace*{-0.2cm}

\noindent where $ \overline{\mathbf{s}} = 1/n \sum_j \mathbf{s}_j$; and $\mathbf{U}_{c_0}, \mathbf{U}_{h_0} \in \mathbb{R}^{2d_h \times d_s}$, $\mathbf{b}_{c_0}, \mathbf{b}_{h_0} \in \mathbb{R}^{d_s}$, with $d_s$ as the number of neural units. These are parameters jointly trained with our entire model.   

\verb"MSIN" incorporates information learnt in the text articles to every step it performs reasoning on the time series in a selective manner. Specifically, at each timestep $\ell$ in the time series sequence\footnote{We re-use $\ell$ to index timestep in a time series sequence, similar to its meaning used with a word sequence in Eq.\eqref{txt_hidd}-\eqref{txt_context_vec}. However, here  $\ell = 1,\ldots,m$.}, \verb"MSIN" searches through the text representative vectors to assign higher probability mass to those that better align with the signal it discovers so far in the time series sequence, captured in the hidden state vector $\mathbf{h}^{ts}_{\ell-1}$. In particular, the attention mass associated with each text representative vector $\mathbf{s}_j$ is computed at $\ell$-th timestep as follows:

\vspace*{-0.3cm}
\begin{align}
a_{\ell,j} &= \tanh(\mathbf{W}_{a}*\mathbf{h}^{ts}_{\ell-1} + \mathbf{U}_a*\mathbf{s}_j + \mathbf{b}_a) \label{eq:U_a} \\
p_{\ell} &= softmax(\mathbf{v}_a^T [a_{\ell,1},a_{\ell,2}, \ldots, a_{\ell,n}])
\end{align}
\vspace*{-0.3cm}

\noindent where $\mathbf{W}_{a}\in \mathbb{R}^{d_s\times d_s}$, $\mathbf{U}_{a} \in \mathbb{R}^{2d_h \times d_s}$, $ \mathbf{b}_a \in \mathbb{R}^{d_s}$ and $\mathbf{v}_a \in \mathbb{R}^n$. The parametric vector $\mathbf{v}_a$ is learnt to transform each alignment vector $a_{\ell,j}$ to a scalar and hence, by passing through the softmax function, $p_{\ell} $ is the probability mass distribution over the text representative sequence. We would like the information from these vectors, scaled proportionally by their probability mass, to immediately impact the learning process over the time series. This is made possible through generating a context vector $\mathbf{v}_\ell$:

\vspace*{-0.5cm}
\begin{align}
\mathbf{v}_\ell &= \frac{1}{2}(\sum_j p_{\ell,j} *\mathbf{s}_j + \mathbf{v}_{\ell-1}) \label{eq:context}
\end{align}
\vspace*{-0.3cm}

\noindent in which $\mathbf{v}_{0}$ is initialized as a zero vector. As designed, \verb"MSIN" constructs the latest context vector as the average information between the current  representation of relevant text article (1st term on the right hand side of Eq.\eqref{eq:context}) and the previous context vector $ \mathbf{v}_{\ell-1}$. By induction, influence of context vectors in the early time steps is fading out as \verb"MSIN" advances in the time series sequence. \verb"MSIN" uses this aggregated vector to regulate the information flow to all its input, forget and output gates:

% \vspace*{-0.5cm}
% \noindent\begin{minipage}[t]{.5\textwidth}
%     \begin{align}
%         \mathbf{i}_\ell &= \sigma (\mathbf{U}_{ix} \mathbf{x}^{ts}_{\ell} + \mathbf{U}_{ih} \mathbf{h}^{ts}_{\ell-1} + \mathbf{U}_{iv} \mathbf{v}_{\ell} + \mathbf{b}_i)  \nonumber \\
%         \mathbf{f}_\ell &= \sigma (\mathbf{U}_{fx} \mathbf{x}^{ts}_{\ell} + \mathbf{U}_{fh} \mathbf{h}^{ts}_{\ell-1} + \mathbf{U}_{fv} \mathbf{v}_{\ell} + \mathbf{b}_f)  \nonumber
%     \end{align}
% \end{minipage}%
% \begin{minipage}[t]{.5\textwidth}
%     \begin{align}
%         \mathbf{o}_\ell &= \sigma (\mathbf{U}_{ox} \mathbf{x}^{ts}_{\ell} + \mathbf{U}_{oh} \mathbf{h}^{ts}_{\ell-1} + \mathbf{U}_{ov} \mathbf{v}_{\ell} + \mathbf{b}_o) \nonumber \\
%         \tilde{\mathbf{c}}^{ts}_\ell &= \tanh (\mathbf{U}_{cx} \mathbf{x}^{ts}_{\ell} + \mathbf{U}_{ch} \mathbf{h}^{ts}_{\ell-1} + \mathbf{U}_{cv} \mathbf{v}_{\ell} + \mathbf{b}_c) \nonumber
%     \end{align} 
% \end{minipage}

\vspace*{-0.3cm}
\begin{align}
\mathbf{i}_\ell &= \sigma (\mathbf{U}_{ix} * \mathbf{x}^{ts}_{\ell} + \mathbf{U}_{ih} * \mathbf{h}^{ts}_{\ell-1} + \mathbf{U}_{iv} * \mathbf{v}_{\ell} + \mathbf{b}_i) \nonumber\\
\mathbf{f}_\ell &= \sigma (\mathbf{U}_{fx} * \mathbf{x}^{ts}_{\ell} + \mathbf{U}_{fh} * \mathbf{h}^{ts}_{\ell-1} + \mathbf{U}_{fv} * \mathbf{v}_{\ell} + \mathbf{b}_f)\nonumber\\
\mathbf{o}_\ell &= \sigma (\mathbf{U}_{ox} * \mathbf{x}^{ts}_{\ell} + \mathbf{U}_{oh} * \mathbf{h}^{ts}_{\ell-1} + \mathbf{U}_{ov} * \mathbf{v}_{\ell} + \mathbf{b}_o)\nonumber
\end{align}
\vspace*{-0.3cm}

\noindent and the candidate cell state:

\vspace*{-0.3cm}
\begin{align}
\tilde{\mathbf{c}}^{ts}_\ell &= \tanh (\mathbf{U}_{cx} * \mathbf{x}^{ts}_{\ell} + \mathbf{U}_{ch} * \mathbf{h}^{ts}_{\ell-1} + \mathbf{U}_{cv} * \mathbf{v}_{\ell} + \mathbf{b}_c)\nonumber
\end{align}
\vspace*{-0.3cm}

\noindent where $\mathbf{U}_{\bullet x} \in \mathbb{R}^{D\times d_s}, \mathbf{U}_{\bullet h} \in \mathbb{R}^{2d_h\times d_s}, \mathbf{U}_{\bullet v} \in \mathbb{R}^{n\times d_s}$ and $\mathbf{b}_{\bullet} \in \mathbb{R}^{d_s}$. Let $\odot$ denote the Hadamard product, the current cell and hidden states are then updated in the following order:

\vspace*{-0.3cm}
\begin{align} 
    \mathbf{c}^{ts}_\ell &= \mathbf{f}_\ell \odot \mathbf{c}^{ts}_{\ell-1} + \mathbf{i}_\ell \odot \tilde{\mathbf{c}}^{ts}, \quad
    \mathbf{h}^{ts}_\ell &= \mathbf{o}^{ts}_\ell \odot \tanh(\mathbf{c}^{ts}_\ell) \nonumber
\end{align}

% \begin{align} 
%     \label{eq:c_ts}
%     \mathbf{c}^{ts}_\ell &= \mathbf{f}_\ell \odot \mathbf{c}^{ts}_{\ell-1} + \mathbf{i}_\ell \odot \tilde{\mathbf{c}}^{ts}\\
%     \label{eq:h_ts}
%     \mathbf{h}^{ts}_\ell &= \mathbf{o}^{ts}_\ell \odot \tanh(\mathbf{c}^{ts}_\ell) 
% \end{align}
\vspace*{-0.1cm}

By tightly integrating the information learnt in the textual modality to every step in modeling the time series, our network distributes burden of work in discovering relevant text articles throughout the course of the series sequence. The selected relevant documents are also immediately exploited to better learn patterns in the time series.

\textit{Output Layer:}
Given representative vectors learnt from the textual domain and the hidden state vectors of the time series, we use a concatenation layer to aggregate them and pass it through an output dense layer. The entire model is trained with the output as the next value in the time series. 

As ablation studies, we consider two variants to our proposed model. First, in order to see the impact of multistep of interrelation between data modalities, we exclude that process from our model, use a conventional LSTM to model the time series and subsequently align its last state with the representative textual vectors to find the information clues. This simplified model has fewer parameters yet the interaction between two data modalities is limited to only the last state of the time series sequence. We name it \verb"LSTMw/o" (LSTM without interaction). Second, excluding the relevant text discovery task, we build a model that treats numerical series and text data independently. Hence, two networks used for two data modality are trained \textit{in parallel} and their outputs are being fused only prior to the last dense layer for making prediction. This model is closely related to the recent work~\cite{akita2016deep}, and we name it  \verb"LSTMpar". Empirically evaluating these models also highlights our model's strength in discovering text articles relevant to the time series.

\vspace*{-0.2cm}
%=========================================================================%
\section{Experiment}                                   \label{Experiment}
%=========================================================================%
%\vspace*{-0.2cm}
%\subsection{Experimental setting}

%\todo{Another column to report P/R by keeping at least 95\% of attention weights (compromise btw 1 and 2 news stories}

\noindent \textbf{Datasets:} 
We analyze dataset consisting of news headlines (text articles) daily collected from Thomson Reuters between years in seven consecutive years from 2006 to  2013~\cite{ReuterData2014}, and a daily stock prices time series of a specific company collected from Yahoo! Finance for the same time period. 
For the results reported below, we form each data sample from the two data modalities as: stock values in the last $m=5$ days (one week trading) as a time series sequence, and all news headlines, ordered by their released time, in the latest 24 hours as a sequence of text documents. With this setting, a trained model aims at discovering a small set of relevant news articles associated with the time series in a daily basis. 

We evaluate models with each of the two company-specific stock time series: (1) Apple (AAPL), and (2) Google (GOOG). Each dataset (text and time series) is split into the training, validation and test sets respectively to the year-periods of 2006-2011, 2012, and 2013. We use the validation set to tune models' parameters, while utilize the test set for an independent evaluation. As mentioned at the beginning, neither time series identity nor fixed keywords have been used to train our model since such an approach requires extensive domain knowledge, is prone to information lost, while also has limited applications. Instead, we let the models automatically learn such identity and keywords by themself. We examine such findings through investigating the text they select to associate with the given time series (more discussion later). 

\noindent \textbf{Baselines:} We name our model \verb"MSIN" as its novel network component. For baseline models, we implement \verb"LSTMw/o", \verb"LSTMpar" as described in the previous section; \verb"GRUtxt" based on~\cite{HAN16}, \verb"CNNtxt"~\cite{kim2014convolutional} that both exploit the textual news modality; and \verb"GRUts" that analyzes the time series. For a conventional machine learning technique, we implemented \verb"SVM"~\cite{weng2017stock} that takes in both time series (as vectors) and the uni-gram for the textual news. 

To emphasize the key contributions, we mainly discuss in the following the results of our model against those from \verb"LSTMw/o" and \verb"GRUtxt" methods in the key task of discovering relevant documents in association with a given time series. This is because they are only the methods that can infer selected documents based on their neural attention mechanism. For comparison results against other techniques and on the prediction task, we report them in the supplementary.

\begin{figure}[t!]
\vspace*{-0.3cm}
\centering
\hspace*{-1.0cm}
\includegraphics[width=9cm]{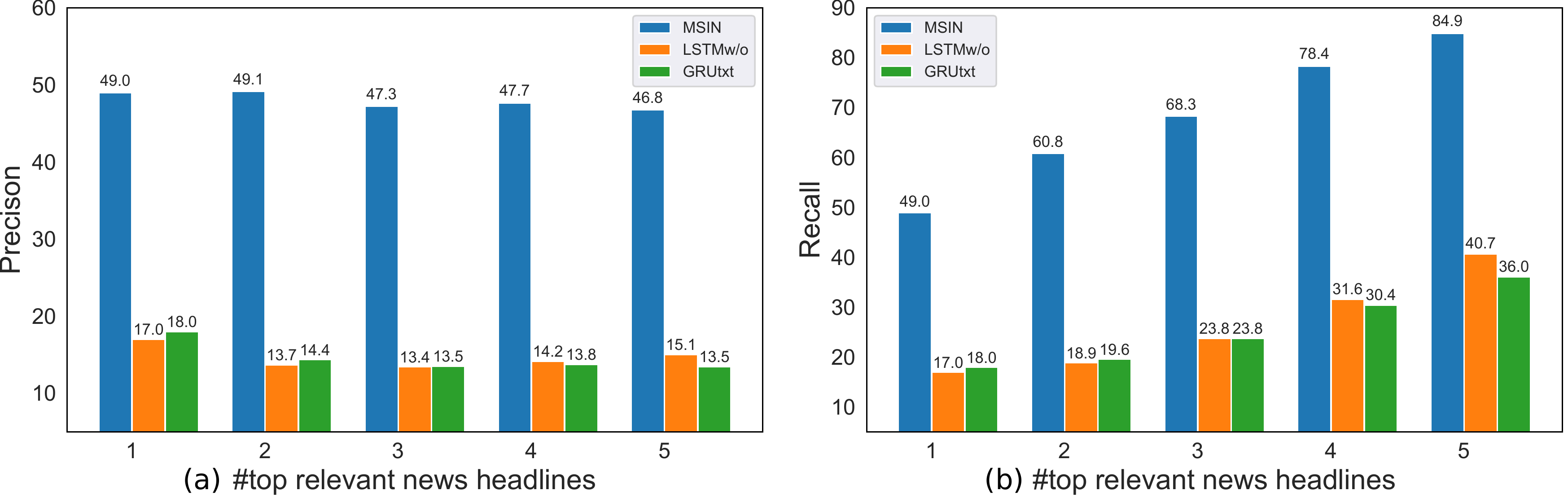}
\caption{(a) Precision and (b) Recall computed w.r.t. ``Apple'' headlines annotated by Reuters.}
\vspace*{-0.5cm}
\label{fig:AAPL_PR}
\end{figure}

\vspace*{-0.1cm}
\subsection{Relevant text articles discovery}  \label{AAPLGOOG-PreRec}
\vspace*{0.1cm}

The Thomson Reuters corpus provides meta-data indicating whether a news article is about a specific company and we use such information as the ground truth relevant news, denoted by GTn. Nonetheless, it is important to emphasize that such information was not used to train our model(s). Neither the identity of time series nor pre-selection of company-specific news articles have been used. Rather, we let \verb"MSIN" learn itself the association between the textual news and the stock series via jointly analyzing the two data modalities simultaneously. \verb"MSIN" is hence \textit{completely data-driven} and is straightfoward to be applied to other corpus such as Bloomberg news source, or other applications like cloud business, where similar ground-truths are not available. 

%  And for a single day, there may have more than one news articles associated with a time series. 
 
\begin{figure}[!t]
\vspace*{-0.3cm}
\centering
\hspace*{-1.0cm}
\includegraphics[width=9cm]{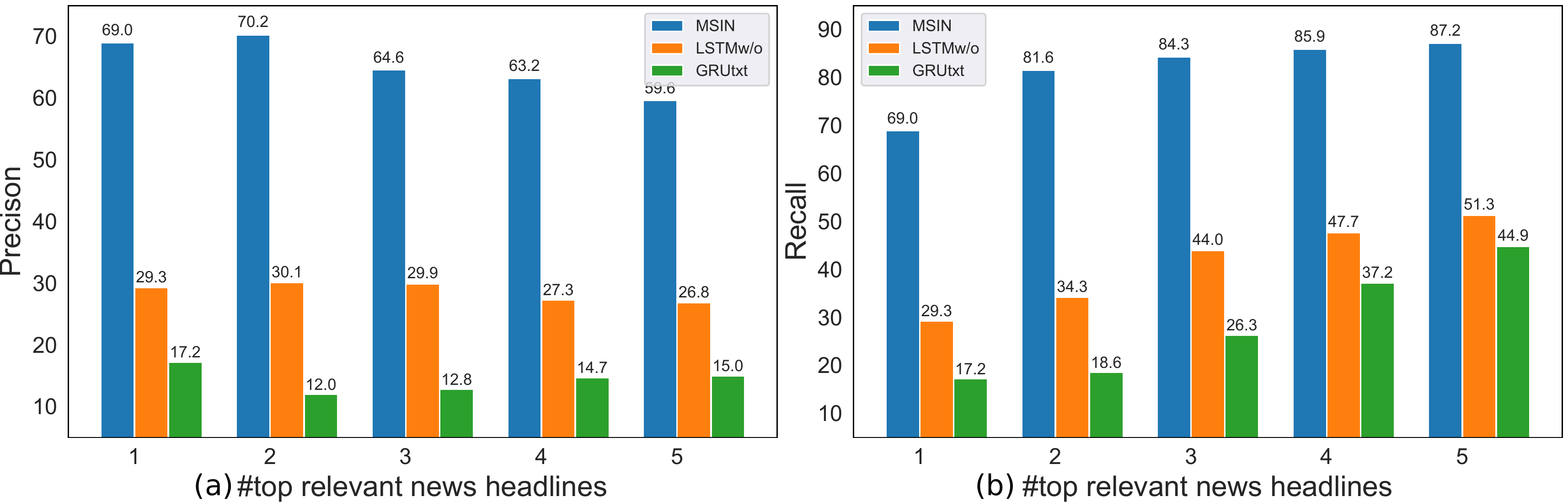}
\caption{(a) Precision and (b) Recall computed w.r.t. ``Google'' headlines annotated by Reuters.}
\vspace*{-0.2cm}
\label{fig:GOOG_PR}
\end{figure}

The GTn headlines allow us to compute the rate of discovering relevant news stories in association with a stock series through the precision and recall metrics. Higher ranking (based on attention mass vector $p_{m}$ as in \verb"MSIN") of these GTn headlines on top of each day signifies better performance of an examined model. Fig.\ref{fig:AAPL_PR}(a-b) and Fig.\ref{fig:GOOG_PR}(a-b) plot these evaluations for the AAPL and GOOG time series respectively, when we vary the number of returned daily top relevant headlines $k$ between 1 and 5 (shown in x-axis). For example, at $k=5$, \verb"MSIN" achieves up to 84.9\% and 87.2\% in recall while retains the precision at 46.8\% and 59.6\% respectively to the GTn sets of AAPL and GOOG. Other settings of $k$ also show that \verb"MSIN"'s performance is far better than the competitive models. Our novel design of fusing two data modalities through multiple-step data interrelations allows our model to effectively discover the complex hidden correlation between the time-varying patterns in the time series and a small set of corresponding information clues in the general textual data. Its precision and recall significantly outperform those of \verb"LSTMw/o" that utilizes only a single step of alignment between the two data modalities, and of \verb"GRUtxt" which solely explores the textual domain with conventional attention mechanism in deep learning~\cite{HAN16}.

%\vspace*{-0.2cm}
\subsection{Explanation on the discovered textual news} \label{exp:explanation}  \label{AAPLGOOG-Explanation}
%\vspace*{0.1cm}

\begin{table}[t!]
\centering
\vspace*{-0.2cm}
  \caption{News headlines from 3 examined days in 2013 (test set). Relevant news headlines discovered by MSIN associated with AAPL stock series are blue-highlighted by setting: accumulated probability mass $\geq 50\%$. (Full set of discovered relevant news are uploaded on our repository due to space constraint)}
  \resizebox{0.5\textwidth}{!}{%
    \begin{tabular}[t]{llll}
    \toprule
    GRUtxt & LSTMw/o & AsyncLSTM & News headlines \\
    \midrule
          &       &       & Date: 2013-01-22 \\
    \colorbox{violet!6}{\makebox[3em]{\strut {0.06}}}  & \colorbox{orange!0}{\makebox[3em]{\strut {0.00}}}  & \colorbox{green!0}{\makebox[3em]{\strut {0.00}}}  &  \colorbox{green!0}{\strut 01.  analysis no respite euro zone long rebalancing slog } \\
    \colorbox{violet!12}{\makebox[3em]{\strut {0.12}}}  & \colorbox{orange!12}{\makebox[3em]{\strut {0.12}}}  & \colorbox{green!15}{\makebox[3em]{\strut {0.15}}}  &  \colorbox{green!0}{\strut 02.  japan government welcome boj ease step towards percent inflation } \\
    \colorbox{violet!6}{\makebox[3em]{\strut {0.06}}}  & \colorbox{orange!4}{\makebox[3em]{\strut {0.04}}}  & \colorbox{green!2}{\makebox[3em]{\strut {0.03}}}  &  \colorbox{green!0}{\strut 03.  japan government panel need achieve budget surplus } \\
    \colorbox{violet!6}{\makebox[3em]{\strut {0.07}}}  & \colorbox{orange!10}{\makebox[3em]{\strut {0.10}}}  & \colorbox{green!21}{\makebox[3em]{\strut {0.21}}}  &  \colorbox{green!21}{\strut 04.  instant view existing home sale fall december } \\
    \colorbox{violet!14}{\makebox[3em]{\strut {0.14}}}  & \colorbox{orange!12}{\makebox[3em]{\strut {0.12}}}  & \colorbox{green!8}{\makebox[3em]{\strut {0.09}}}  &  \colorbox{green!0}{\strut 05.  german exporter fear devaluation round boj move } \\
    \colorbox{violet!7}{\makebox[3em]{\strut {0.07}}}  & \colorbox{orange!5}{\makebox[3em]{\strut {0.05}}}  & \colorbox{green!2}{\makebox[3em]{\strut {0.03}}}  &  \colorbox{green!0}{\strut 06.  bank japan yet revive economy } \\
    \colorbox{violet!7}{\makebox[3em]{\strut {0.07}}}  & \colorbox{orange!6}{\makebox[3em]{\strut {0.06}}}  & \colorbox{green!3}{\makebox[3em]{\strut {0.04}}}  &  \colorbox{green!0}{\strut 07.  home resale fall housing recovery still track } \\
    \colorbox{violet!8}{\makebox[3em]{\strut {0.08}}}  & \colorbox{orange!12}{\makebox[3em]{\strut {0.12}}}  & \colorbox{green!1}{\makebox[3em]{\strut {0.02}}}  &  \colorbox{green!0}{\strut 08.  instant view google put up well than expect quarterly number } \\
    \colorbox{violet!7}{\makebox[3em]{\strut {0.08}}}  & \colorbox{orange!11}{\makebox[3em]{\strut {0.11}}}  & \colorbox{green!2}{\makebox[3em]{\strut {0.03}}}  &  \colorbox{green!0}{\strut 09.  bank japan buy asset sp set new five year high } \\
    \colorbox{violet!12}{\makebox[3em]{\strut {0.12}}}  & \colorbox{orange!6}{\makebox[3em]{\strut {0.06}}}  & \colorbox{green!0}{\makebox[3em]{\strut {0.01}}}  &  \colorbox{green!0}{\strut 10.  google fourth quarter result shine ad rate decline slows } \\
    \colorbox{violet!12}{\makebox[3em]{\strut {0.12}}}  & \colorbox{orange!3}{\makebox[3em]{\strut {0.03}}}  & \colorbox{green!0}{\makebox[3em]{\strut {0.00}}}  &  \colorbox{green!0}{\strut 11.  banks commodity stock lift sp five year high } \\
    \colorbox{violet!7}{\makebox[3em]{\strut {0.07}}}  & \colorbox{orange!2}{\makebox[3em]{\strut {0.02}}}  & \colorbox{green!0}{\makebox[3em]{\strut {0.01}}}  &  \colorbox{green!0}{\strut 12.  google fourth quarter result shine ad rate decline slows } \\
    \colorbox{violet!6}{\makebox[3em]{\strut {0.07}}}  & \colorbox{orange!13}{\makebox[3em]{\strut {0.14}}}  & \colorbox{green!39}{\makebox[3em]{\strut {0.40}}}  &  \colorbox{green!39}{\strut 13.  steve jobs threaten patent suit no hire policy filing } \\
    \midrule
          &       &       & Date: 2013-08-14 \\
    \colorbox{violet!11}{\makebox[3em]{\strut {0.11}}}  & \colorbox{orange!3}{\makebox[3em]{\strut {0.03}}}  & \colorbox{green!0}{\makebox[3em]{\strut {0.00}}}  &  \colorbox{green!0}{\strut 01.  france exit recession beat second quarter gdp forecast } \\
    \colorbox{violet!9}{\makebox[3em]{\strut {0.10}}}  & \colorbox{orange!14}{\makebox[3em]{\strut {0.14}}}  & \colorbox{green!0}{\makebox[3em]{\strut {0.00}}}  &  \colorbox{green!0}{\strut 02.  euro zone performance suggests recovery sight european rehn } \\
    \colorbox{violet!14}{\makebox[3em]{\strut {0.14}}}  & \colorbox{orange!3}{\makebox[3em]{\strut {0.03}}}  & \colorbox{green!0}{\makebox[3em]{\strut {0.00}}}  &  \colorbox{green!0}{\strut 03.  germany france haul euro zone recession } \\
    \colorbox{violet!11}{\makebox[3em]{\strut {0.11}}}  & \colorbox{orange!6}{\makebox[3em]{\strut {0.07}}}  & \colorbox{green!0}{\makebox[3em]{\strut {0.00}}}  &  \colorbox{green!0}{\strut 04.  yellen see likely next fed chair despite summers chatter reuters poll } \\
    \colorbox{violet!11}{\makebox[3em]{\strut {0.11}}}  & \colorbox{orange!12}{\makebox[3em]{\strut {0.12}}}  & \colorbox{green!0}{\makebox[3em]{\strut {0.00}}}  &  \colorbox{green!0}{\strut 05.  us modest recovery fed cut back qe next month reuters poll } \\
    \colorbox{violet!4}{\makebox[3em]{\strut {0.04}}}  & \colorbox{orange!6}{\makebox[3em]{\strut {0.06}}}  & \colorbox{green!2}{\makebox[3em]{\strut {0.03}}}  &  \colorbox{green!0}{\strut 06.  j.c. penney share spike report sale improve august } \\
    \colorbox{violet!10}{\makebox[3em]{\strut {0.10}}}  & \colorbox{orange!5}{\makebox[3em]{\strut {0.06}}}  & \colorbox{green!0}{\makebox[3em]{\strut {0.00}}}  &  \colorbox{green!0}{\strut 07.  wallstreet end down fed uncertainty data boost europe } \\
    \colorbox{violet!6}{\makebox[3em]{\strut {0.06}}}  & \colorbox{orange!2}{\makebox[3em]{\strut {0.02}}}  & \colorbox{green!1}{\makebox[3em]{\strut {0.02}}}  &  \colorbox{green!0}{\strut 08.  analysis balloon google experiment web access } \\
    \colorbox{violet!5}{\makebox[3em]{\strut {0.05}}}  & \colorbox{orange!11}{\makebox[3em]{\strut {0.11}}}  & \colorbox{green!0}{\makebox[3em]{\strut {0.01}}}  &  \colorbox{green!0}{\strut 09.  wallstreet fall uncertainty fed bond buying } \\
    \colorbox{violet!6}{\makebox[3em]{\strut {0.06}}}  & \colorbox{orange!9}{\makebox[3em]{\strut {0.09}}}  & \colorbox{green!75}{\makebox[3em]{\strut {0.76}}}  &  \colorbox{green!75}{\strut 10.  apple face possible may trial e book damage } \\
    \colorbox{violet!3}{\makebox[3em]{\strut {0.04}}}  & \colorbox{orange!3}{\makebox[3em]{\strut {0.03}}}  & \colorbox{green!17}{\makebox[3em]{\strut {0.17}}}  &  \colorbox{green!0}{\strut 11.  japan government spokesman no pm abe corporate tax cut } \\
    \midrule
          &       &       & Date: 2013-09-06 \\
    \colorbox{violet!2}{\makebox[3em]{\strut {0.03}}}  & \colorbox{orange!3}{\makebox[3em]{\strut {0.04}}}  & \colorbox{green!20}{\makebox[3em]{\strut {0.21}}}  &  \colorbox{green!20}{\strut 01.  china unicom telecom sell late iphone shortly us launch } \\
    \colorbox{violet!3}{\makebox[3em]{\strut {0.03}}}  & \colorbox{orange!4}{\makebox[3em]{\strut {0.05}}}  & \colorbox{green!1}{\makebox[3em]{\strut {0.01}}}  &  \colorbox{green!0}{\strut 02.  spain industrial output fall month july } \\
    \colorbox{violet!4}{\makebox[3em]{\strut {0.05}}}  & \colorbox{orange!4}{\makebox[3em]{\strut {0.05}}}  & \colorbox{green!0}{\makebox[3em]{\strut {0.00}}}  &  \colorbox{green!0}{\strut 03.  french consumer confidence trade point improve outlook } \\
    \colorbox{violet!8}{\makebox[3em]{\strut {0.08}}}  & \colorbox{orange!2}{\makebox[3em]{\strut {0.02}}}  & \colorbox{green!0}{\makebox[3em]{\strut {0.00}}}  &  \colorbox{green!0}{\strut 04.  uk industrial output flat july trade deficit widens sharply } \\
    \colorbox{violet!3}{\makebox[3em]{\strut {0.04}}}  & \colorbox{orange!2}{\makebox[3em]{\strut {0.02}}}  & \colorbox{green!0}{\makebox[3em]{\strut {0.00}}}  &  \colorbox{green!0}{\strut 05.  boj kuroda room policy response tax hike hurt economy minute } \\
    \colorbox{violet!5}{\makebox[3em]{\strut {0.05}}}  & \colorbox{orange!12}{\makebox[3em]{\strut {0.12}}}  & \colorbox{green!5}{\makebox[3em]{\strut {0.05}}}  &  \colorbox{green!0}{\strut 06.  wind down market street funding amid regulatory pressure } \\
    \colorbox{violet!9}{\makebox[3em]{\strut {0.09}}}  & \colorbox{orange!3}{\makebox[3em]{\strut {0.03}}}  & \colorbox{green!1}{\makebox[3em]{\strut {0.01}}}  &  \colorbox{green!0}{\strut 07.  china able cope fed policy taper central bank head } \\
    \colorbox{violet!4}{\makebox[3em]{\strut {0.04}}}  & \colorbox{orange!4}{\makebox[3em]{\strut {0.05}}}  & \colorbox{green!2}{\makebox[3em]{\strut {0.02}}}  &  \colorbox{green!0}{\strut 08.  instant view us august nonfarm payroll rise } \\
    \colorbox{violet!4}{\makebox[3em]{\strut {0.05}}}  & \colorbox{orange!4}{\makebox[3em]{\strut {0.04}}}  & \colorbox{green!3}{\makebox[3em]{\strut {0.04}}}  &  \colorbox{green!0}{\strut 09.  analysis fed shift syria crisis trading strategy } \\
    \colorbox{violet!2}{\makebox[3em]{\strut {0.03}}}  & \colorbox{orange!4}{\makebox[3em]{\strut {0.04}}}  & \colorbox{green!0}{\makebox[3em]{\strut {0.01}}}  &  \colorbox{green!0}{\strut 10.  factbox three thing learn us job report } \\
    \colorbox{violet!4}{\makebox[3em]{\strut {0.05}}}  & \colorbox{orange!4}{\makebox[3em]{\strut {0.04}}}  & \colorbox{green!9}{\makebox[3em]{\strut {0.09}}}  &  \colorbox{green!0}{\strut 11.  g20 say economy recover but no end crisis yet } \\
    \colorbox{violet!3}{\makebox[3em]{\strut {0.03}}}  & \colorbox{orange!3}{\makebox[3em]{\strut {0.04}}}  & \colorbox{green!3}{\makebox[3em]{\strut {0.04}}}  &  \colorbox{green!0}{\strut 12.  us regulator talk european energy price probe } \\
    \colorbox{violet!3}{\makebox[3em]{\strut {0.04}}}  & \colorbox{orange!3}{\makebox[3em]{\strut {0.04}}}  & \colorbox{green!0}{\makebox[3em]{\strut {0.01}}}  &  \colorbox{green!0}{\strut 13.  wallstreet flat job data syria worry spur caution } \\
    \colorbox{violet!3}{\makebox[3em]{\strut {0.04}}}  & \colorbox{orange!3}{\makebox[3em]{\strut {0.03}}}  & \colorbox{green!0}{\makebox[3em]{\strut {0.00}}}  &  \colorbox{green!0}{\strut 14.  job growth disappoints offer note fed } \\
    \colorbox{violet!4}{\makebox[3em]{\strut {0.04}}}  & \colorbox{orange!4}{\makebox[3em]{\strut {0.05}}}  & \colorbox{green!2}{\makebox[3em]{\strut {0.02}}}  &  \colorbox{green!0}{\strut 15.  bond yield dollar fall us job data } \\
    \colorbox{violet!4}{\makebox[3em]{\strut {0.05}}}  & \colorbox{orange!5}{\makebox[3em]{\strut {0.05}}}  & \colorbox{green!37}{\makebox[3em]{\strut {0.37}}}  &  \colorbox{green!37}{\strut 16.  apple hit us injunction e books antitrust case } \\
    \colorbox{violet!5}{\makebox[3em]{\strut {0.06}}}  & \colorbox{orange!12}{\makebox[3em]{\strut {0.12}}}  & \colorbox{green!0}{\makebox[3em]{\strut {0.01}}}  &  \colorbox{green!0}{\strut 17.  wallstreet week ahead markets could turn choppy fed syria risk mount } \\
    \colorbox{violet!3}{\makebox[3em]{\strut {0.04}}}  & \colorbox{orange!3}{\makebox[3em]{\strut {0.04}}}  & \colorbox{green!1}{\makebox[3em]{\strut {0.01}}}  &  \colorbox{green!0}{\strut 18.  china buy giant kazakh oilfield billion } \\
    \colorbox{violet!5}{\makebox[3em]{\strut {0.05}}}  & \colorbox{orange!4}{\makebox[3em]{\strut {0.04}}}  & \colorbox{green!1}{\makebox[3em]{\strut {0.01}}}  &  \colorbox{green!1}{\strut 19.  italy wo n't block foreign takeover economy minister } \\
    \colorbox{violet!4}{\makebox[3em]{\strut {0.05}}}  & \colorbox{orange!2}{\makebox[3em]{\strut {0.03}}}  & \colorbox{green!0}{\makebox[3em]{\strut {0.00}}}  &  \colorbox{green!0}{\strut 20.  china august export beat forecast point stabilization } \\
    \colorbox{violet!12}{\makebox[3em]{\strut {0.12}}}  & \colorbox{orange!06}{\makebox[3em]{\strut {0.06}}}  & \colorbox{green!0}{\makebox[3em]{\strut {0.01}}}  &  \colorbox{green!0}{\strut 21.  wallstreet week ahead markets could turn choppy fed syria risk mount } \\
    \colorbox{violet!4}{\makebox[3em]{\strut {0.04}}}  & \colorbox{orange!2}{\makebox[3em]{\strut {0.03}}}  & \colorbox{green!1}{\makebox[3em]{\strut {0.02}}}  &  \colorbox{green!0}{\strut 22.  india inc policymakers shape up ship } \\
    \colorbox{violet!2}{\makebox[3em]{\strut {0.02}}}  & \colorbox{orange!2}{\makebox[3em]{\strut {0.02}}}  & \colorbox{green!3}{\makebox[3em]{\strut {0.04}}}  &  \colorbox{green!0}{\strut 23.  cost lack indonesia economy } \\
    \colorbox{violet!2}{\makebox[3em]{\strut {0.03}}}  & \colorbox{orange!1}{\makebox[3em]{\strut {0.02}}}  & \colorbox{green!1}{\makebox[3em]{\strut {0.01}}}  &  \colorbox{green!0}{\strut 24.  mexico proposes new tax regime pemex } \\
    \bottomrule
    \end{tabular}%
  }
  \label{tab:AAPL-attn}%
  \vspace*{-0.2cm}
\end{table}

As concrete examples for qualitative evaluation, we show in Tables~\ref{tab:AAPL-attn} all news headlines from three specific examined days, along with those discovered by our model (highlighted in green and by setting their cumulated probability mass of attention $\geq 50\%$) as relevant news when it was trained with the AAPL series, and evaluated performance on the independent test dataset. As clearly seen on date 2013-01-22, \verb"MSIN" gives high attention mass to \textit{``steve jobs threatened patent suit to enforce no-hire policy"} though none of words mention the Apple company. Likewise, on 2013-09-06, \emph{``china unicom, telecom to sell latest iphone shortly after u.s. launch''} received the 2nd highest probability mass (21\%), in addition to the 1st one \emph{``apple hit with u.s. injunction in e-books antitrust case''} (37\%). Although ground truth news headlines (often containing company name keyword) have been used to quantitatively evaluate \verb"MSIN"'s performance (Fig.\ref{fig:AAPL_PR} and \ref{fig:GOOG_PR}), we believe that these uncovered news headlines demonstrate the success and potential of \verb"MSIN", as our model is capable to unearth the news content that never explicitly mention company names. We observe similar performance of \verb"MSIN" when it is trained with the GOOG stock series, using the same news corpus as presented in Table~\ref{tab:GOOG-attn}. The results present the probability mass of attention of \verb"MSIN" on three days selected from the test dataset. Note that date 2013-08-14 are deliberately shown in both Tables~\ref{tab:AAPL-attn} and \ref{tab:GOOG-attn} to demonstrate that the set of relevant news are clearly dependent on which time series has been used to train our model along with the general text corpus. Their relevancy to each of two time series is obvious and intuitively interpretable.

\begin{table}[t!]
 \centering
  \caption{News headlines from 3 examined days in 2013 (test set). Relevant news headlines discovered by MSIN associated with GOOG stock series are blue-highlighted by setting: accumulated probability mass $\geq 50\%$.}
  \resizebox{0.48\textwidth}{!}{%
    \begin{tabular}[t]{llll}
    \toprule
    GRUtxt & LSTMw/o & AsyncLSTM & News headlines \\
    \midrule
          &       &       & Date: 2013-01-09 \\
    \colorbox{violet!23}{\makebox[3em]{\strut {0.23}}}  & \colorbox{orange!34}{\makebox[3em]{\strut {0.34}}}  & \colorbox{green!0}{\makebox[3em]{\strut {0.00}}}  &  \colorbox{green!0}{\strut 01.  short sellers circle stock confidence waver } \\
    \colorbox{violet!26}{\makebox[3em]{\strut {0.26}}}  & \colorbox{orange!52}{\makebox[3em]{\strut {0.52}}}  & \colorbox{green!87}{\makebox[3em]{\strut {0.87}}}  &  \colorbox{green!87}{\strut 02.  google drop key patent claim microsoft } \\
    \colorbox{violet!21}{\makebox[3em]{\strut {0.21}}}  & \colorbox{orange!10}{\makebox[3em]{\strut {0.10}}}  & \colorbox{green!0}{\makebox[3em]{\strut {0.00}}}  &  \colorbox{green!0}{\strut 03.  alcoa result lift share dollar up vs. yen } \\
    \colorbox{violet!10}{\makebox[3em]{\strut {0.10}}}  & \colorbox{orange!3}{\makebox[3em]{\strut {0.03}}}  & \colorbox{green!0}{\makebox[3em]{\strut {0.00}}}  &  \colorbox{green!0}{\strut 04.  wallstreet rise alcoa report earnings } \\
    \midrule
          &       &       & Date: 2013-08-14 \\
    \colorbox{violet!10}{\makebox[3em]{\strut {0.11}}}  & \colorbox{orange!10}{\makebox[3em]{\strut {0.10}}}  & \colorbox{green!0}{\makebox[3em]{\strut {0.00}}}  &  \colorbox{green!0}{\strut 01.  france exit recession beat second quarter gdp forecast } \\
    \colorbox{violet!9}{\makebox[3em]{\strut {0.10}}}  & \colorbox{orange!4}{\makebox[3em]{\strut {0.05}}}  & \colorbox{green!0}{\makebox[3em]{\strut {0.00}}}  &  \colorbox{green!0}{\strut 02.  euro zone performance suggests recovery sight european rehn } \\
    \colorbox{violet!11}{\makebox[3em]{\strut {0.11}}}  & \colorbox{orange!6}{\makebox[3em]{\strut {0.07}}}  & \colorbox{green!0}{\makebox[3em]{\strut {0.00}}}  &  \colorbox{green!0}{\strut 03.  germany france haul euro zone recession } \\
    \colorbox{violet!10}{\makebox[3em]{\strut {0.10}}}  & \colorbox{orange!10}{\makebox[3em]{\strut {0.11}}}  & \colorbox{green!0}{\makebox[3em]{\strut {0.00}}}  &  \colorbox{green!0}{\strut 04.  yellen see likely next fed chair despite summers chatter reuters poll } \\
    \colorbox{violet!11}{\makebox[3em]{\strut {0.12}}}  & \colorbox{orange!8}{\makebox[3em]{\strut {0.08}}}  & \colorbox{green!0}{\makebox[3em]{\strut {0.00}}}  &  \colorbox{green!0}{\strut 05.  us modest recovery fed cut back qe next month reuters poll } \\
    \colorbox{violet!5}{\makebox[3em]{\strut {0.05}}}  & \colorbox{orange!3}{\makebox[3em]{\strut {0.04}}}  & \colorbox{green!1}{\makebox[3em]{\strut {0.02}}}  &  \colorbox{green!0}{\strut 06.  j.c. penney share spike report sale improve august } \\
    \colorbox{violet!9}{\makebox[3em]{\strut {0.10}}}  & \colorbox{orange!5}{\makebox[3em]{\strut {0.05}}}  & \colorbox{green!0}{\makebox[3em]{\strut {0.00}}}  &  \colorbox{green!0}{\strut 07.  wallstreet end down fed uncertainty data boost europe } \\
    \colorbox{violet!7}{\makebox[3em]{\strut {0.08}}}  & \colorbox{orange!17}{\makebox[3em]{\strut {0.18}}}  & \colorbox{green!79}{\makebox[3em]{\strut {0.79}}}  &  \colorbox{green!79}{\strut 08.  analysis balloon google experiment web access } \\
    \colorbox{violet!8}{\makebox[3em]{\strut {0.09}}}  & \colorbox{orange!8}{\makebox[3em]{\strut {0.08}}}  & \colorbox{green!0}{\makebox[3em]{\strut {0.00}}}  &  \colorbox{green!0}{\strut 09.  wallstreet fall uncertainty fed bond buying } \\
    \colorbox{violet!5}{\makebox[3em]{\strut {0.06}}}  & \colorbox{orange!16}{\makebox[3em]{\strut {0.16}}}  & \colorbox{green!5}{\makebox[3em]{\strut {0.05}}}  &  \colorbox{green!0}{\strut 10.  apple face possible may trial e book damage } \\
    \colorbox{violet!3}{\makebox[3em]{\strut {0.04}}}  & \colorbox{orange!7}{\makebox[3em]{\strut {0.08}}}  & \colorbox{green!2}{\makebox[3em]{\strut {0.02}}}  &  \colorbox{green!0}{\strut 11.  japan government spokesman no pm abe corporate tax cut } \\
    \midrule
          &       &       & Date: 2013-08-29 \\
    \colorbox{violet!8}{\makebox[3em]{\strut {0.08}}}  & \colorbox{orange!9}{\makebox[3em]{\strut {0.09}}}  & \colorbox{green!1}{\makebox[3em]{\strut {0.01}}}  &  \colorbox{green!0}{\strut 01.  india pm likely make statement economy friday } \\
    \colorbox{violet!7}{\makebox[3em]{\strut {0.07}}}  & \colorbox{orange!11}{\makebox[3em]{\strut {0.11}}}  & \colorbox{green!0}{\makebox[3em]{\strut {0.01}}}  &  \colorbox{green!0}{\strut 02.  boj warns emerge market may see outflow } \\
    \colorbox{violet!6}{\makebox[3em]{\strut {0.06}}}  & \colorbox{orange!9}{\makebox[3em]{\strut {0.09}}}  & \colorbox{green!3}{\makebox[3em]{\strut {0.03}}}  &  \colorbox{green!0}{\strut 03.  european rehn say lender step up assessment greece next month } \\
    \colorbox{violet!3}{\makebox[3em]{\strut {0.03}}}  & \colorbox{orange!6}{\makebox[3em]{\strut {0.06}}}  & \colorbox{green!20}{\makebox[3em]{\strut {0.20}}}  &  \colorbox{green!20}{\strut 04.  china environment min suspends approval cnpc } \\
    \colorbox{violet!4}{\makebox[3em]{\strut {0.05}}}  & \colorbox{orange!12}{\makebox[3em]{\strut {0.12}}}  & \colorbox{green!7}{\makebox[3em]{\strut {0.08}}}  &  \colorbox{green!0}{\strut 05.  italy still meet target scrap property tax say rehn } \\
    \colorbox{violet!4}{\makebox[3em]{\strut {0.04}}}  & \colorbox{orange!5}{\makebox[3em]{\strut {0.05}}}  & \colorbox{green!1}{\makebox[3em]{\strut {0.01}}}  &  \colorbox{green!1}{\strut 06.  spain economic slump longer than thought but ease } \\
    \colorbox{violet!5}{\makebox[3em]{\strut {0.06}}}  & \colorbox{orange!4}{\makebox[3em]{\strut {0.04}}}  & \colorbox{green!1}{\makebox[3em]{\strut {0.02}}}  &  \colorbox{green!0}{\strut 07.  india central bank consider gold trade minister } \\
    \colorbox{violet!5}{\makebox[3em]{\strut {0.06}}}  & \colorbox{orange!2}{\makebox[3em]{\strut {0.02}}}  & \colorbox{green!2}{\makebox[3em]{\strut {0.02}}}  &  \colorbox{green!0}{\strut 08.  rupee fall front slow indian economy } \\
    \colorbox{violet!5}{\makebox[3em]{\strut {0.05}}}  & \colorbox{orange!1}{\makebox[3em]{\strut {0.02}}}  & \colorbox{green!1}{\makebox[3em]{\strut {0.02}}}  &  \colorbox{green!0}{\strut 09.  india rupee bounce record low pm may address economy } \\
    \colorbox{violet!3}{\makebox[3em]{\strut {0.04}}}  & \colorbox{orange!4}{\makebox[3em]{\strut {0.05}}}  & \colorbox{green!0}{\makebox[3em]{\strut {0.01}}}  &  \colorbox{green!0}{\strut 10.  exclusive india might buy gold ease rupee crisis } \\
    \colorbox{violet!6}{\makebox[3em]{\strut {0.06}}}  & \colorbox{orange!4}{\makebox[3em]{\strut {0.05}}}  & \colorbox{green!1}{\makebox[3em]{\strut {0.02}}}  &  \colorbox{green!0}{\strut 11.  spain recession longer than thought but close end } \\
    \colorbox{violet!5}{\makebox[3em]{\strut {0.05}}}  & \colorbox{orange!5}{\makebox[3em]{\strut {0.05}}}  & \colorbox{green!1}{\makebox[3em]{\strut {0.01}}}  &  \colorbox{green!0}{\strut 12.  india finance minister asks bank ensure credit flow industry } \\
    \colorbox{violet!6}{\makebox[3em]{\strut {0.06}}}  & \colorbox{orange!8}{\makebox[3em]{\strut {0.08}}}  & \colorbox{green!0}{\makebox[3em]{\strut {0.01}}}  &  \colorbox{green!0}{\strut 13.  easing stimulus weigh oil next year reuters poll } \\
    \colorbox{violet!3}{\makebox[3em]{\strut {0.03}}}  & \colorbox{orange!4}{\makebox[3em]{\strut {0.05}}}  & \colorbox{green!0}{\makebox[3em]{\strut {0.01}}}  &  \colorbox{green!0}{\strut 14.  exclusive india might buy gold ease rupee crisis } \\
    \colorbox{violet!3}{\makebox[3em]{\strut {0.04}}}  & \colorbox{orange!2}{\makebox[3em]{\strut {0.02}}}  & \colorbox{green!2}{\makebox[3em]{\strut {0.02}}}  &  \colorbox{green!0}{\strut 15.  india rupee bounce record low government seek solution } \\
    \colorbox{violet!3}{\makebox[3em]{\strut {0.03}}}  & \colorbox{orange!2}{\makebox[3em]{\strut {0.02}}}  & \colorbox{green!37}{\makebox[3em]{\strut {0.37}}}  &  \colorbox{green!37}{\strut 16.  china google power global drive } \\
    \colorbox{violet!09}{\makebox[3em]{\strut {0.09}}}  & \colorbox{orange!2}{\makebox[3em]{\strut {0.03}}}  & \colorbox{green!0}{\makebox[3em]{\strut {0.00}}}  &  \colorbox{green!0}{\strut 17.  gdp growth beat forecast may boost case fed move } \\
    \colorbox{violet!8}{\makebox[3em]{\strut {0.08}}}  & \colorbox{orange!1}{\makebox[3em]{\strut {0.01}}}  & \colorbox{green!0}{\makebox[3em]{\strut {0.00}}}  &  \colorbox{green!0}{\strut 18.  wallstreet rise economy but syria concern limit gain } \\
    \colorbox{violet!5}{\makebox[3em]{\strut {0.06}}}  & \colorbox{orange!1}{\makebox[3em]{\strut {0.01}}}  & \colorbox{green!3}{\makebox[3em]{\strut {0.03}}}  &  \colorbox{green!0}{\strut 19.  oil dip syria action uncertain dollar rise data } \\
    \colorbox{violet!1}{\makebox[3em]{\strut {0.02}}}  & \colorbox{orange!0}{\makebox[3em]{\strut {0.00}}}  & \colorbox{green!0}{\makebox[3em]{\strut {0.01}}}  &  \colorbox{green!0}{\strut 20.  boe carney say uncertainty rbs future end } \\
    \bottomrule
    \end{tabular}%
  }
  \label{tab:GOOG-attn}%
\end{table}

%\vspace*{-0.2cm}
%=========================================================================%
\section{Related work}                                   \label{Related work}

Large number of single-modality studies analyzing either time series data or unstructured text documents have been proposed in the literature. Some of these studies are based on classical statistical methods~\cite{wu2016grey,michell2018stock} or neural networks~\cite{qiu2016application,zhong2017forecasting}.  Recent studies from the financial domain explore both time series of asset prices and the news articles~\cite{schumaker12yahoonews,weng2017stock,akita2016deep} which are related to our work. Typically, these studies attempt to transform text from financial news into various numerical forms including news sentiment, subjective polarity, n-grams and combine them with stock data. These handcrafted features
require extensive pre-processing and are also extracted independently from the time series data. A more recent model~\cite{akita2016deep} relies on RNNs that enable it to model stock series in their natural form and later merge them with the vector representation of textual news prior to making the market prediction. The goal of all these studies remains to improve prediction accuracy but not for the time series explanation purpose. They hence lacks the capability of providing relevant interpretation for time series based on the textual information. 

%We argue that the impact of lacking explainability would be significant in analyzing and understanding financial stock markets nowadays. This is particularly true given the fact the amount of available information has dramatically increased these years, causing neither institutional traders nor investors can follow and consider all available information. Thus, an automated forecasting system with the capability of interrelating information from multiple data sources and  instantly providing explanation is important in enhancing our understanding regarding the key factors behind the observable data.

Our work is also related to multi-modal deep learning studies~\cite{MM19Survey} which generally can be classified into three categories: early, late, and hybrid (in-between),  depending on how and at which level data from multiple modalities are fused together. In early fusion~\cite{valada2016deep,zadeh2016multimodal}, multi-modal data sources are concatenated into a single feature vector prior to being used as inputs for a learning model, while in late fusion, data is aggregated from the outputs of multiple models, each trained on a separate modality and fused later based on aggregation rules such as, averaged-fusion~\cite{nojavanasghari2016deep}, tensor products~\cite{zadeh2017tensor}, or a meta-model like gated memory~\cite{zadeh2018memory}. The hybrid (in-between) fusion is the trade-off paradigm, which allows the data to be aggregated at multiple scales, yet often requiring synchronization among data modalities, such as in the synchronized gesture recognition~\cite{neverova2015moddrop,rajagopalan2016extending}. Our model is related to the third category; yet, we deal with asynchronous multimodals of numerical time series and unstructured text and relax the constraints on the time-step synchronization between modalities. More significantly,  we perform data fusion through multiple steps and at the low-level features which strengthens our model in learning associated patterns across data modalities.

\vspace*{-0.2cm}
%=========================================================================%
\section{Conclusion}                                   \label{conclusion}
%=========================================================================%
\vspace*{0.1cm}

Jointly learning both numerical time series and unstructured textual data is an important research endeavor to enhance the understanding of time series performance. In this work, we presented a novel neural model that is capable of discovering the top relevant textual information associated with a given time series. In dealing with the complexity of relationship between two data modalities, with different data sampling rates and lengths, we develop \verb"MSIN" that allows the direct incorporation of information learnt in the textual modality to every time step modeling on the behavior of time series, considerably leveraging their mutual influence through time. Through this multi-step data interrelation, \verb"MSIN" can learn to focus on a small subset of textual data that best aligns with the time series. We demonstrate the performance of our model
in the financial domain using time series of two stock prices, which are trained along with a corpus of news headlines collected from Thompson Reuters. Our \verb"MSIN" model discovers relevant news stories to the stocks that do not even explicitly mention the company name, but include highly relevant events that may influence or reflect their performance in the market.

%\footnotesize
%\scriptsize
\bibliography{paper} 
\bibliographystyle{aaai}
%\pagebreak
%\pagebreak
\appendix
%=========================================================================%
%\section{General architecture of our neural network model}                             \label{architecture}
%=========================================================================%

%\begin{figure}[!h]
%\centering
%\includegraphics[width=9cm]{OverallModel.pdf}
%\caption{The high-level architecture of our model that jointly learns both numerical time series and unstructured textual news to discover top most relevant news associated with a given time series and offer them as an intuitive explanation for the timely performance of the time series.}
%\label{OverallModel}
%\end{figure}

%=========================================================================%
\section{Supplementary}                            
%=========================================================================%
%=========================================================================%
\subsection{Supp. Thomson Reuters Data}                             \label{textData}
\begin{figure}[!h]
\vspace*{-0.1cm}
\centering
%\hspace*{-1.0cm}
\includegraphics[width=7cm]{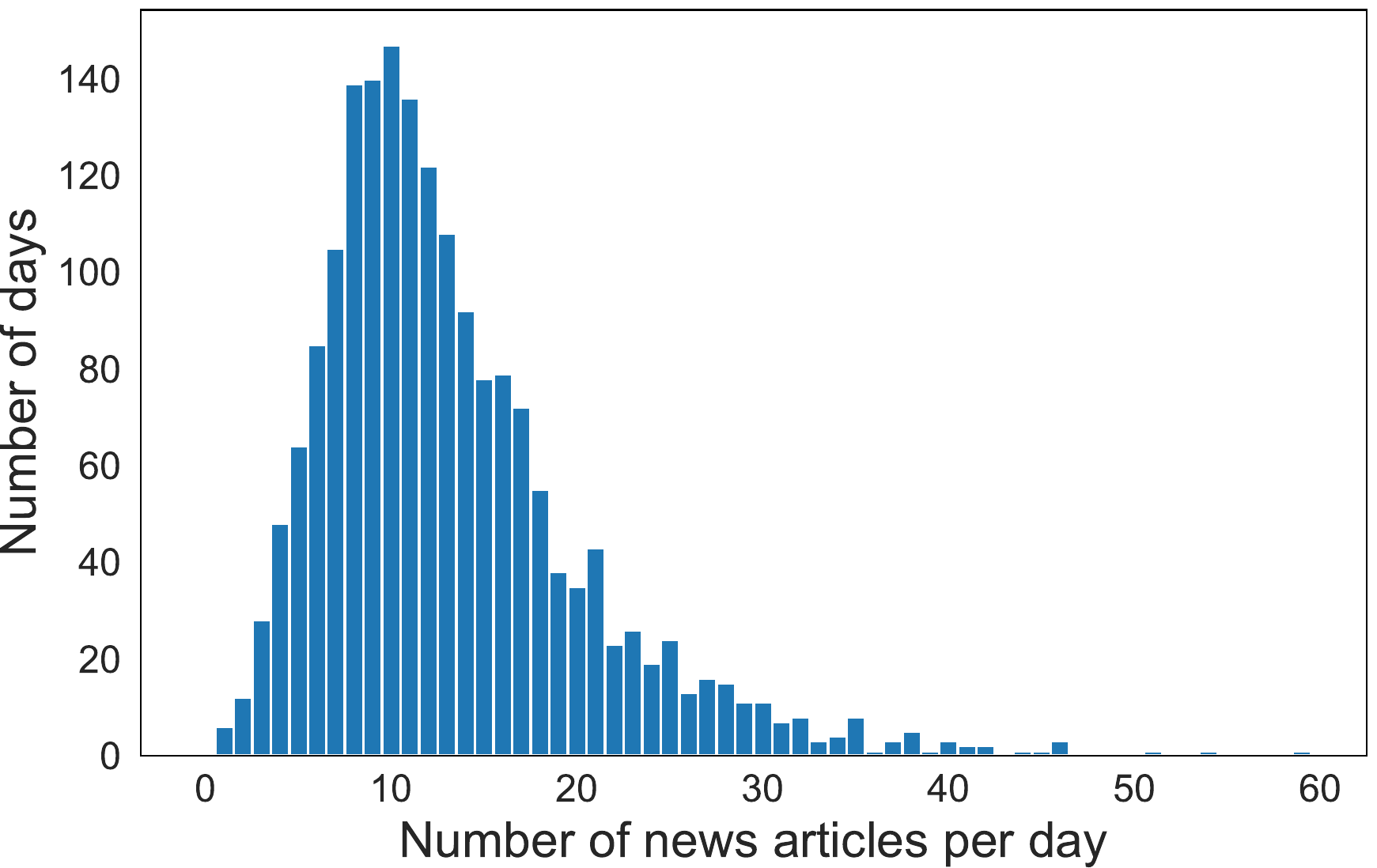}
\caption{Distribution of numbers of news headlines released per day from the Thomson Reuters news corpus.}
\vspace*{-0.1cm}
\label{NumhlinesPerDay}
\end{figure}

%=========================================================================%
Fig.~\ref{NumhlinesPerDay} plots the distribution of numbers of news headlines released per day by Reuters. As a skewed distribution (mode being around 12 headlines per day), we limit all models to explore up to 25 news articles per day. This setting retains more than 95\% of the total number of headlines in the original textual corpus. For a small portion of days having exceptionally large numbers of daily news articles (those located to the far-right of the distribution shown in Fig.~\ref{NumhlinesPerDay} ), we keep the last 25 news headlines on each of these days. We use NLTK toolkit (available at \url{https://www.nltk.org/}) for tex preprocessing, keep the vocabulary size at 5000 unique words, and remove ones that are not found in the 400k words of the GloVe. 

%=========================================================================%
\subsection{Supp. Statistics on Ground Truth News Headlines}                             \label{GTn}
%=========================================================================%

\begin{table}[h]
  \centering
  %\vspace*{-0.7cm}
  \centering
  \caption{Statistics on  ``ground truth" news (GTn) headlines on test dataset associated with company-specific time series. \%GTn shows percentage of GTn over all news headlines. \%GTd shows percentage of days having at least one GTn. GTn/GTd shows average percentage of GTn on GTd.  Max\#GTn shows the maximum number of GTn headlines in a single day.}
  \vspace*{0.4cm}
   %\resizebox{0.4\textwidth}{!}{%
    \begin{tabular}{crrrr}
    \toprule
    Dataset & \%GTn & \%GTd & GTn/GTd & Max\#GTn \\
    \midrule
    AAPL  & 8\%   & 45\%  & 15\%  & 9 \\
    GOOG  & 3.5\% & 26\%  & 12\%  & 5 \\
    \bottomrule
    \end{tabular}%
  % }
  \label{tab:stat}%
\end{table}

%=========================================================================%
\subsection{Supp. Parameters Setting}                             \label{param}
%=========================================================================%

 We set the parameter ranges for our models and its counterparts as follows: The number of layers' neural units $d_s, d_h \in \{16,32,64,128\}$, regularization $L_1, L_2 \in \{0.1,0.05,0.01,0.005, 0.001\}$,  dropout rate $\in\{0.0, 0.1, 0.2, 0.4\}$, and the number of timesteps (sequent length) for stock time series $m \in\{3,5,7,10\}$. We choose 50 as the word embedding dimension with the usage of the pre-trained GloVe model~\cite{GLOVE}. For \verb"GRUtxt", we follows the model presented in~\cite{HAN16} using two Bi-LSTMs along with a self-attention layer applied at the sentence level. With CNNtxt, we implement an architecture with three kernel filters of sizes of $\{2,3,4\}$, using max pooling as recommended in~\cite{kim2014convolutional,CNNtxt}. We use two layers of LSTMs as encoder-decoder for the \verb"GRUts", while SVM is used with the linear kernel and C is tuned from the log range Our model and the baselines were implemented and trained on a machine with two Tesla K80 GPUs, running Tensorflow 1.3.0 as backend. 
We performed the random search using the validation set to tune hyperparameters. 
 
 The final values we used for AAPL dataset is $d_s, d_h = 64, m = 5, L_1 = 0.01, L_2 = 0.2$, GOOG is $d_s = 32 , d_h = 64, m = 5, L_1 = 0.001, L_2 = 0.1$, both with dropout of 0.2 and using pre-trained GloVe embedding, For S\&P500, the values are $d_s = 64 , d_h = 96, m = 10, L_1 = 0.001, L_2 = 0.005$ with dropout of 0.1 and initializing with GloVe for word embedding.

%=========================================================================%
\subsection{Supp. Evaluation Metrics}                             \label{PR}
%=========================================================================%

In empirical evaluation of our models and its counterparts, we have used the precision and recall measurements. While their calculation with respect to predicting class labels remains as usual, their computation with respect to the ground truth news headlines (GTn) is slightly changed, being adaptive to the setting $k$ as the number of top relevant headlines to be returned, as reported in Fig.~\ref{fig:AAPL_PR}(a-b), Fig.\ref{fig:GOOG_PR}(a-b). This is because some days have as few as only one GTn, other days may have as many as 9 (in AAPL) or 5 (in GOOG) as shown in Appendix~\ref{GTn}, while the cardinality $k$ is fixed in computing precision and recall. Specifically, we compute the $\text{Pre}@k$ and $\text{Rec}@k$ (reported in Fig.~\ref{fig:AAPL_PR}(a-b), Fig.\ref{fig:GOOG_PR}(a-b)) averaged from all days having at least one ground truth headline as follows:

\begin{align} 
   \label{eq:Pre-k}
\text{Pre}@k &= \frac{1}{\text{GTd}}\sum_{i=1}^\text{GTd} \frac{TP_i[:k]}{TP_i[:k] + FP_i[:k]}\\
\label{eq:Rec-k} 
\text{Rec}@k &= \frac{1}{\text{GTd}}\sum_{i=1}^\text{GTd} \frac{TP_i[:k]}{\min(k,|\text{GTn}_i[:k]|)}
\end{align}

\noindent in which $TP_i[:k], FP_i[:k]$ are respectively the numbers of true positive and false positive headlines at day $i$, and $\text{GTn}_i[:k]$ denotes the number of ground truth news headlines up to $k$ cardinality on day $i$. The denominator in $\text{Rec}@k$ ensures that the number of false negative headlines will not be beyond either the cardinality $k$ or the actual number of ground truth headlines on day $i$.

\begin{table}[t!]
\centering
 \caption{News headlines from three days selected from the test set, along with the attention mass (annotated in the numeric columns) by: GRUtxt, LSTMw/o and MSIN. Colored headlines are relevant news outputted by our model based on setting the accumulated probability mass $\geq 50\%$.}
  \vspace*{0.1cm}
  \resizebox{0.5\textwidth}{!}{%
    \begin{tabular}{cccl}
    \toprule
    GRUtxt & LSTMw/o & MSIN & News headlines \\
    \midrule
%    -/-   & -/+   & -/-   &  Date: 2013-02-03 \\
       &    &    &  Date: 2013-02-03 \\
     \colorbox{violet!0}{\makebox[3em]{\strut{0.00}}} &  \colorbox{orange!53}{\makebox[3em]{\strut{0.53}}} &  \colorbox{green!0}{\makebox[3em]{\strut{0.00}}} & 01. china service slow uptick highlight recovery \\
     \colorbox{violet!8}{\makebox[3em]{\strut{0.08}}} &  \colorbox{orange!10}{\makebox[3em]{\strut{0.10}}} &  \colorbox{green!0}{\makebox[3em]{\strut{0.00}}} & 02. japan finance minister weak yen result not goal anti deflation policy \\
     \colorbox{violet!2}{\makebox[3em]{\strut{0.02}}} &  \colorbox{orange!1}{\makebox[3em]{\strut{0.01}}} &  \colorbox{green!18}{\makebox[3em]{\strut{0.18}}} & 03. france track meet percent growth target minister \\
     \colorbox{violet!5}{\makebox[3em]{\strut{0.05}}} &  \colorbox{orange!16}{\makebox[3em]{\strut{0.16}}} &  \colorbox{green!0}{\makebox[3em]{\strut{0.00}}} & 04. watch central banker say not \\
     \colorbox{violet!12}{\makebox[3em]{\strut{0.12}}} &  \colorbox{orange!1}{\makebox[3em]{\strut{0.01}}} &  \colorbox{green!8}{\makebox[3em]{\strut{0.08}}} & 05. us small business borrowing rise december but barely \\
     \colorbox{violet!28}{\makebox[3em]{\strut{0.28}}} &  \colorbox{orange!8}{\makebox[3em]{\strut{0.08}}} &  \colorbox{green!7}{\makebox[3em]{\strut{0.07}}} & 06. gauge us business investment plan edge low \\
     \colorbox{violet!15}{\makebox[3em]{\strut{0.15}}} &  \colorbox{orange!7}{\makebox[3em]{\strut{0.07}}} &  \colorbox{green!39}{\makebox[3em]{\strut{0.39}}} & \colorbox{green!39}{\strut 07. global share euro fall sharply renew euro zone fear} \\
     \colorbox{violet!30}{\makebox[3em]{\strut{0.30}}} &  \colorbox{orange!3}{\makebox[3em]{\strut{0.03}}} &  \colorbox{green!21}{\makebox[3em]{\strut{0.21}}} & \colorbox{green!21}{\strut 08. sp post bad day since november hill share sink} \\
\midrule
%    -/-   & -/-   & -/-   &   Date: 2013-04-26 \\
       &    &    &   Date: 2013-04-26 \\
     \colorbox{violet!0}{\makebox[3em]{\strut{0.00}}} &  \colorbox{orange!0}{\makebox[3em]{\strut{0.00}}} &  \colorbox{green!0}{\makebox[3em]{\strut{0.00}}} & 01. japan growth strategy fiscal plan g8 \\
     \colorbox{violet!0}{\makebox[3em]{\strut{0.00}}} &  \colorbox{orange!1}{\makebox[3em]{\strut{0.01}}} &  \colorbox{green!0}{\makebox[3em]{\strut{0.00}}} & 02. boj credibility test division emerge over inflation target \\
     \colorbox{violet!28}{\makebox[3em]{\strut{0.28}}} &  \colorbox{orange!4}{\makebox[3em]{\strut{0.04}}} &  \colorbox{green!1}{\makebox[3em]{\strut{0.01}}} & 03. consumer sentiment wane april \\
     \colorbox{violet!0}{\makebox[3em]{\strut{0.00}}} &  \colorbox{orange!1}{\makebox[3em]{\strut{0.01}}} &  \colorbox{green!5}{\makebox[3em]{\strut{0.05}}} & 04. economic growth gauge year high last week ecri \\
     \colorbox{violet!0}{\makebox[3em]{\strut{0.00}}} &  \colorbox{orange!1}{\makebox[3em]{\strut{0.01}}} &  \colorbox{green!0}{\makebox[3em]{\strut{0.00}}} & 05. microsoft get upper hand first google patent trial \\
     \colorbox{violet!2}{\makebox[3em]{\strut{0.02}}} &  \colorbox{orange!1}{\makebox[3em]{\strut{0.01}}} &  \colorbox{green!1}{\makebox[3em]{\strut{0.01}}} & 06. growth fall short forecast weakness ahead \\
     \colorbox{violet!22}{\makebox[3em]{\strut{0.22}}} &  \colorbox{orange!9}{\makebox[3em]{\strut{0.09}}} &  \colorbox{green!14}{\makebox[3em]{\strut{0.14}}} & 07. dollar fall yen bond yield decline \\
     \colorbox{violet!28}{\makebox[3em]{\strut{0.28}}} &  \colorbox{orange!80}{\makebox[3em]{\strut{0.80}}} &  \colorbox{green!73}{\makebox[3em]{\strut{0.73}}} & \colorbox{green!73}{\strut 08. wallstreet dip gdp but finish week high} \\
     \colorbox{violet!19}{\makebox[3em]{\strut{0.19}}} &  \colorbox{orange!4}{\makebox[3em]{\strut{0.04}}} &  \colorbox{green!5}{\makebox[3em]{\strut{0.05}}} & 09. wallstreet week ahead central bank data steer investor \\
    \midrule
%    +/+   & +/+   & +/+   &  Date:  2013-10-18 \\
       &    &    &  Date:  2013-10-18 \\
     \colorbox{violet!0}{\makebox[3em]{\strut{0.00}}} &  \colorbox{orange!2}{\makebox[3em]{\strut{0.02}}} &  \colorbox{green!0}{\makebox[3em]{\strut{0.00}}} & 01. china economy show sign slow september stats bureau \\
     \colorbox{violet!0}{\makebox[3em]{\strut{0.00}}} &  \colorbox{orange!1}{\makebox[3em]{\strut{0.01}}} &  \colorbox{green!0}{\makebox[3em]{\strut{0.00}}} & 02. china share up comfort data but still weekly loss \\
     \colorbox{violet!0}{\makebox[3em]{\strut{0.00}}} &  \colorbox{orange!8}{\makebox[3em]{\strut{0.08}}} &  \colorbox{green!1}{\makebox[3em]{\strut{0.01}}} & 03. china third quarter gdp growth fast year but outlook dim \\
     \colorbox{violet!2}{\makebox[3em]{\strut{0.02}}} &  \colorbox{orange!26}{\makebox[3em]{\strut{0.26}}} &  \colorbox{green!7}{\makebox[3em]{\strut{0.07}}} & 04. google third quarter beat ad volume grows stock flirt level \\
     \colorbox{violet!1}{\makebox[3em]{\strut{0.01}}} &  \colorbox{orange!11}{\makebox[3em]{\strut{0.11}}} &  \colorbox{green!3}{\makebox[3em]{\strut{0.03}}} & 05. schlumberger baker top estimate global drilling \\
     \colorbox{violet!26}{\makebox[3em]{\strut{0.26}}} &  \colorbox{orange!2}{\makebox[3em]{\strut{0.02}}} &  \colorbox{green!1}{\makebox[3em]{\strut{0.01}}} & 06. us data boon computer driven trading \\
     \colorbox{violet!0}{\makebox[3em]{\strut{0.00}}} &  \colorbox{orange!0}{\makebox[3em]{\strut{0.00}}} &  \colorbox{green!0}{\makebox[3em]{\strut{0.00}}} & 07. fed release delayed report industrial output october \\
     \colorbox{violet!0}{\makebox[3em]{\strut{0.00}}} &  \colorbox{orange!0}{\makebox[3em]{\strut{0.00}}} &  \colorbox{green!0}{\makebox[3em]{\strut{0.00}}} & 08. data delay add us inflation bond sale \\
     \colorbox{violet!20}{\makebox[3em]{\strut{0.20}}} &  \colorbox{orange!10}{\makebox[3em]{\strut{0.10}}} &  \colorbox{green!7}{\makebox[3em]{\strut{0.07}}} & 09. stock split wane may follow google 1,000 \\
     \colorbox{violet!6}{\makebox[3em]{\strut{0.06}}} &  \colorbox{orange!3}{\makebox[3em]{\strut{0.03}}} &  \colorbox{green!41}{\makebox[3em]{\strut{0.41}}} & \colorbox{green!41}{\strut 10. global stock gain dollar fall fed see stay course} \\
     \colorbox{violet!44}{\makebox[3em]{\strut{0.44}}} &  \colorbox{orange!31}{\makebox[3em]{\strut{0.31}}} &  \colorbox{green!39}{\makebox[3em]{\strut{0.39}}} & \colorbox{green!39}{\strut 11. sp break record google stock top earnings} \\
     \colorbox{violet!1}{\makebox[3em]{\strut{0.01}}} &  \colorbox{orange!3}{\makebox[3em]{\strut{0.03}}} &  \colorbox{green!0}{\makebox[3em]{\strut{0.00}}} & 12. google share break mobile pay \\
    \bottomrule
    \end{tabular}%
    }
  \label{tab:sp500visual}%
\end{table}

%=========================================================================%
\section{Supp. Experiment on S\&P500 Time Series}                             \label{SP}
%=========================================================================%

We show in Table~\ref{tab:sp500visual} the headline news in three days selected from the test dataset, along with the probability mass of attention (on valid news headlines) of our \verb"MSIN" model, its variant \verb"LSTMw/o" and \verb"GRUtxt" respectively in 3rd-1st columns. 
As observed, our model MSIN places more probability mass on the headlines whose contents directly report the market performance, while zeros out mass on headlines that report company-events or local markets. Its attention mass is also more condense, with clear focus on a small set of top relevant news, as compared to those of \verb"LSTMw/o" and \verb"GRUtxt" that both spread out on multiple headlines. The headlines highlighted in Table~\ref{tab:sp500visual} are those outputted by our model by setting their cumulated probability mass of at least 50\%.

The focus of our work is on discovering relevant text articles in association with the numerical time series. As using the global market time series for experiment, nevertheless, we would like to test the impact of selected relevant text articles discovered by our model in forecasting the overall market movement in the subsequent day. For this task, we further compare its performances against other baseline models, including \verb"CNNtxt"~\cite{kim2014convolutional} that both exploit the textual news modality; \verb"GRUts" that analyzes the time series; and \verb"LSTMpar", closely related to~\cite{akita2016deep}, which trains \text{in parallel} two LSTM networks, each on one data modality, then fuses them together prior to the output layer. Moreover, for a conventional machine learning technique, we implemented \verb"SVM"~\cite{weng2017stock} that takes in both time series (as vectors) and the uni-gram for the textual news. 

\vspace*{-0.2cm}
\begin{table}[htbp]
  \centering
  \vspace*{-0.3cm}
  \caption{\small{Performance of all models on forecasting market movement. Precision and recall are reported w.r.t. two prediction of up and down market respectively.}}
  \vspace*{0.2cm}
    \resizebox{0.35\textwidth}{!}{%
   \begin{tabular}{cccc}
    \toprule
    Models & Acc.  & Pre.  & Rec. \\
    \midrule
    LSTMw/o & 55.8  & \textbf{63.9}/48.2 & 56.1/\textbf{56.4} \\
    LSTMpar & 53.6  & 57.8/43.5 & 70.1/31.3 \\
    GRUts & 52.8  & 58.1/43.4 & 65.7/33.6 \\
    GRUtxt & 55.3  & 60.1/45.1 & 70.9/33.3 \\
    CNNtxt & 55.9  & 61.4/46.9 & 64.1/45.3 \\
    SVM   & 53.1  & 58.1/41.2 & \textbf{71.7}/27.4 \\
    MSIN & \textbf{56.4}  & 63.3/\textbf{50.1} & 56.9/56.3 \\
    \bottomrule
    \end{tabular}%
   }
  \label{tab:sp500Accuracy}%
\end{table}%

\vspace*{0.2cm}

Table~\ref{tab:sp500Accuracy} reports the forecasting accuracy, precision and recall of all models. Two values in each entry of precision and recall columns are respectively reported for the forecasting of up and down market on the next day respectively. As observed, there is not much difference in the prediction accuracy of \verb"GRUtxt" and \verb"CNNtxt" though they analyze the textual news with different network topologies. Without exploiting the temporal order in the time series and the semantic dependencies of textual news, \verb"SVM"'s performance seems less comparable to its counterpart neural networks. Out of all models, our \verb"MSIN" achieves higher prediction accuracy. Its solution also converges to a more balanced classification boundary as reflected in the precision and recall measures. 

\vspace*{-0.2cm}
\begin{table}[htbp]
 \vspace*{-0.3cm}
  \centering
  \caption{\small{Performance of all models on forecasting market movement as varying the daily latest time up to which news headlines are collected.}}
  \vspace*{0.2cm}
   \resizebox{0.35\textwidth}{!}{%
    \begin{tabular}{crrrr}
    \toprule
    Models & 9:00  & 15:40 & 19:00 & 23:59 \\
    \midrule
    LSTMw/o & 56.1  & 57.2  & 67.7  & 77.6 \\
    LSTMpar & 54.5  & 56.1  & 65.9  & 76.1 \\
    GRUtxt & 56.9  & 58.1  & 68.3  & 78.2 \\
    CNNtxt & 56.8  & \textbf{60.1}  & 69.1  & 78.4 \\
    SVM   & 54.3  & 55.8  & 61.2  & 75.9 \\
    MSIN & \textbf{57.1}  & 59.8  & \textbf{69.6}  & \textbf{79.3} \\
    \bottomrule
    \end{tabular}%
   }
  \label{tab:sp500TimeVarying}%
\end{table}%

\vspace*{0.2cm}
In order to further observe the impact of textual news on the models' classification accuracy, we vary the latest time, up to which the daily news headlines are collected, to 9:00 (before market open), 15:40 (before market close), 19:00, and 23:59 on the same day the market performance is predicted. These evaluations are reported in Table \ref{tab:sp500TimeVarying}. A clear trend is seen that, the prediction accuracy is higher as the news articles are collected closer to the market closing time, which confirms the indicative information embedded in the textual news toward predicting the S\&P500 performance. 
While most of neural models tend to perform feature engineering better than that of \verb"SVM", only our \verb"MSIN" can further offer better interpretation
due to its join-training of both evolving time series and the news articles. It is noted that evaluations at the time stamp of 19:00 (or 23:59) will shift a model from a predictive system to a purely explanatory one.

\end{document}